\documentclass[10pt]{article} 
 \usepackage[accepted]{tmlr}

\usepackage{microtype}

\usepackage{hyperref}
\usepackage{url}
\usepackage{siunitx}
\usepackage{booktabs}
\usepackage{algorithm}
\usepackage{algorithmic}

\usepackage{arydshln}

\definecolor{darkblue}{rgb}{0, 0, 0.5}
\definecolor{DarkGreen}{RGB}{1,50,32}

\hypersetup{colorlinks=true, citecolor=darkblue, linkcolor=darkblue, urlcolor=darkblue}

\usepackage{microtype}

\usepackage{graphicx}
\usepackage{wrapfig,lipsum,booktabs}
\usepackage{multirow}
\definecolor{LimeGreen}{rgb}{0.15,0.65,0.15}
\definecolor{Gray}{gray}{0.6}
\definecolor{Black}{gray}{0.0}
\usepackage{fdsymbol}  

\usepackage{caption}
\usepackage{subcaption}
\usepackage{listings}
\definecolor{codegreen}{rgb}{0,0.6,0}
\definecolor{codegray}{rgb}{0.5,0.5,0.5}
\definecolor{codepink}{RGB}{252, 142, 172}
\definecolor{codepurple}{rgb}{0.58,0,0.82}
\definecolor{backcolour}{RGB}{245,245,245}
\lstdefinestyle{mystyle}{
    backgroundcolor=\color{backcolour},   
    commentstyle=\color{magenta},
    keywordstyle=\color{blue},
    numberstyle=\tiny\color{codegray},
    stringstyle=\color{codepurple},
    basicstyle=\fontfamily{\ttdefault}\footnotesize,
    breakatwhitespace=false,         
    breaklines=true,                 
    keepspaces=true,    
    frame=single,
    numbersep=5pt,                  
    showspaces=false,                
    showstringspaces=false,
    showtabs=false,                  
    tabsize=2,
    classoffset=1, 
    keywordstyle=\color{violet},
    classoffset=0,
}
\lstdefinelanguage{JavaScript}{
  keywords={typeof, new, true, false, catch, function, return, null, catch, switch, var, if, in, while, do, else, case, break},
  keywordstyle=\color{blue}\bfseries,
  ndkeywords={class, export, boolean, throw, implements, import, this},
  ndkeywordstyle=\color{darkgray}\bfseries,
  identifierstyle=\color{black},
  sensitive=false,
  comment=[l]{//},
  morecomment=[s]{/*}{*/},
  commentstyle=\color{purple}\ttfamily,
  stringstyle=\color{red}\ttfamily,
  morestring=[b]',
  morestring=[b]''
}

\title{LitLLMs, LLMs for Literature Review: Are we there yet?}

\author{\name Shubham Agarwal *  \\
  \addr ServiceNow Research, Mila - Quebec AI Institute, HEC Montreal \\ \\
  \name Gaurav Sahu *  \\
  \addr ServiceNow Research, University of Waterloo \\ \\
  \name Abhay Puri *  \\
  \addr ServiceNow Research \\ \\
  \name Issam H. Laradji \\
  \addr ServiceNow Research, University of British Columbia \\ \\
  \name Krishnamurthy DJ Dvijotham  \\
  \addr ServiceNow Research \\ \\
  \name Jason Stanley  \\
  \addr ServiceNow Research \\ \\
  \name Laurent Charlin  \\
  \addr Mila - Quebec AI Institute, HEC Montreal, Canada CIFAR AI Chair \\ \\
  \name Christopher Pal  \\
  \addr ServiceNow Research, Polytechnique Montreal, Mila - Quebec AI Institute, Canada CIFAR AI Chair \\ \\
  \textsuperscript{*} Equal contribution
}

\def\month{12}  
\def\year{2024} 
\def\openreview{\url{https://openreview.net/forum?id=heeJqQXKg7}} 

\begin{document}

\maketitle

\begin{abstract}
Literature reviews are an essential component of scientific research, but they remain time-intensive and challenging to write, especially due to the recent influx of research papers. This paper explores the zero-shot abilities of recent Large Language Models (LLMs) in assisting with the writing of literature reviews based on an abstract.
We decompose the task into two components: (1) Retrieving related works given a query abstract and  (2) Writing a literature review based on the retrieved results. We analyze how effective LLMs are for both components.
For retrieval, we introduce a novel two-step search strategy that first uses an LLM to extract meaningful keywords from the abstract of a paper and then retrieves potentially relevant papers by querying an external knowledge base. Additionally, we study a prompting-based re-ranking mechanism with attribution and show that re-ranking doubles the normalized recall compared to naive search methods while providing insights into the LLM's decision-making process.
In the generation phase, we propose a two-step approach that first outlines a plan for the review and then executes steps in the plan to generate the actual review.
To evaluate different LLM-based literature review methods, we create test sets from arXiv papers using a protocol designed for rolling use with newly released LLMs to avoid test set contamination in zero-shot evaluations.
We release this evaluation protocol to promote additional research and development in this regard.
Our empirical results suggest that LLMs show promising potential for writing literature reviews when the task is decomposed into smaller components of retrieval and planning.
Particularly, we find that combining keyword-based and document-embedding-based search improves precision and recall during retrieval by 10\% and 30\%, respectively, compared to using either of the methods in isolation.
Further, we demonstrate that our planning-based approach achieves higher-quality reviews by minimizing hallucinated references in the generated review by 18-26\% compared to existing simpler LLM-based generation methods.
Our project page including a demonstration system and toolkit can be accessed here: \url{https://litllm.github.io}.
\end{abstract}

\section{Introduction}


Writing a literature review---finding, citing, and contextualizing relevant prior work---is a fundamental scientific writing requirement. When writing manuscripts, scientists must situate their proposed ideas within the existing literature.
Writing a good literature review is a complex task which can be broken down into two broad sub-tasks: \textbf{1)} Finding relevant papers and \textbf{2)} Generating a related work section to discuss the proposed research given prior works.
This challenge is further amplified in fields such as machine learning, where thousands of relevant papers appear every month on arXiv alone.\footnote{E.g.\ over 4,000 ML papers were submitted to arXiv in October 2024: \url{https://arxiv.org/list/cs.LG/2024-10}}
We explore the utility and potential of large language models (LLMs), in combination with retrieval mechanisms, to assist in generating comprehensive literature reviews for scientific papers.
Specifically, we investigate using LLMs to generate a paper's related work section based on its abstract. We use the term abstract loosely, not necessarily to refer to the actual abstract of the paper but rather to a textual passage that captures a concise summary of the paper's key contributions and scope. Using the abstract as input allows our system to target the central ideas of the paper without requiring the complete manuscript, which is often continuously evolving in the early stages of writing. While our experiments focus on using the abstract, our framework is designed to be flexible. It can use the entire manuscript as it evolves, albeit at a higher computational cost and the need to use models that support longer context windows. This approach provides valuable early-stage insights for authors seeking preliminary references to shape their work, with the capacity to seamlessly incorporate additional information as the manuscript develops.

\begin{figure}
\centering
\includegraphics[scale=0.6]{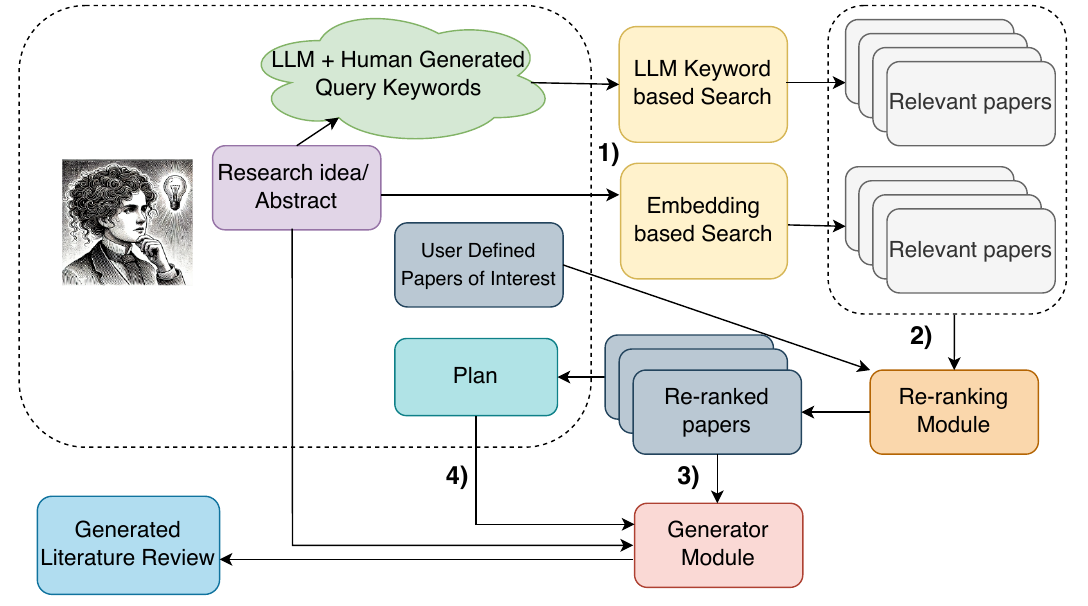}
\caption{\small A schematic diagram of our framework,  where: 1) Relevant prior work is retrieved using keyword and embedding-based search. 2) LLMs re-rank results to find the most relevant prior work. 3) Based on these papers and the user abstract or idea summary, an LLM generates a literature review, 4) optionally controlled by a sentence plan.}
\label{fig:flow}
\end{figure}

The architecture of our framework is illustrated in Figure \ref{fig:flow}, where we further decompose each of the two above tasks into two subtasks.
In this work:
\textbf{1)} We introduce an LLM-based approach to retrieve relevant papers, where we first extract the keywords from an abstract or research idea paragraph using an LLM and then feed these keywords to a keyword-based search tool --- we experiment with Google search and Semantic Scholar. We optionally also transform the abstract or idea into an embedding and use an embedding-based search procedure. 
\textbf{2)} We then employ a prompting-based approach to rank the set of retrieved candidate papers based on their relevance to the query abstract, also requiring the LLM to attribute the relevance to specific excerpts in the candidate papers. We explore multiple re-ranking and aggregation strategies.  
\textbf{3)} To generate the literature review, we select top-$k$ papers from the ranked list and prompt the LLM to generate the related work section based on the query abstract and the abstracts of the selected papers.
\textbf{4)} Additionally, we examine the effectiveness of providing a writing plan to the LLM that specifies which papers to cite at various points in the literature review. These plans can be generated entirely by the LLM, by the user, or a combination of the two. These plans serve as an intermediate representation giving the user more control over the organizational structure of the literature review. 

The complete framework involves multiple innovations, where we use LLMs in multiple ways, namely for generating search queries, re-ranking search results, and attribution. We summarize the main contributions of our work as follows:
\begin{itemize}
  \item To answer the key question that our paper poses, we present a data collection protocol and multiple instances of using it to collect arXiv papers. Critically, our protocol is based on using the most recent month of arXiv papers in a rolling manner with the goal of avoiding test-set contamination when evaluating the most recent LLMs for literature review-related tasks. We then use this protocol to perform extensive retrieval and literature review generation experiments. We release both our datasets and our code to the community.
  \item We propose a novel LLM-based pipeline for the task of interactive literature review writing, which we decompose into two distinct components: retrieval and generation.
  This also facilitates more controlled studies investigating alternative LLM-based approaches for these sub-tasks. Our experiments focus on evaluating fully automated variants of these sub-tasks, but our framing of the problem and our proposed solutions are easily integrated into scenarios where human users interact with systems that assist them. To the best of our knowledge, our decomposition of the problem into sub-tasks, along with our proposed solutions to them, and our framing of this assistive scenario is novel.  
  \item We make multiple contributions that improve the retrieval phase of the two-step process. First, we propose and evaluate a novel strategy that first uses an LLM to extract meaningful keywords from the abstract of a paper and then queries different types of external sources to retrieve potentially relevant papers. Second, we compare and combine the aforementioned LLM-generated keyword search techniques with document embedding-based retrieval methods.  
  Third, we propose and examine a wide variety of search re-ranking techniques. Among these, we propose a novel prompting for attribution approach, which we find to empirically improve the relevance of retrieved literature while also improving reliability, providing insights and improving transparency in the decision-making process of LLMs when used to rerank results. Going even further, we examine debate-prone LLMs for aggregating and re-ranking keyword search and embedding search results. 
  Our experiments show that combining these ideas improves precision and normalized recall by 10\% and 30\%, respectively, compared to standard retrieval methods.
  \item For text generation, we propose and examine a plan-based retrieval augmented approach to writing literature reviews. By using a plan and conditioning on retrieved context, we provide a user greater control over generated content, and our experiments show that this approach can improve the quality of the generated literature reviews substantially. We evaluate the approach using automated metrics and human assessments and show that our method generates higher-quality reviews as measured by ROUGE scores and as assessed by human evaluation. Our approach also reduces hallucinations by 18--26\%.
\end{itemize}

\section{Related Work}
\label{sec:related-work}



We decompose the literature review task into two key sub-tasks: identifying relevant papers and generating the final related work section. This decomposition enhances the likelihood that the LLMs can effectively accomplish the task. We now discuss the relevant literature pertinent to both aspects of the process.

\subsection{Ranking and Retrieval}
Traditional methods for information retrieval rely on techniques like TF-IDF and BM25 to identify documents that are semantically similar to a given query. More recently, dense vector representations obtained through models like Sentence-BERT~\citep{reimers-sbert} have been shown to improve retrieval accuracy by encoding both query and documents into an embedding space where semantic similarity can be readily computed. The initial retrieval stage often results in a large set of candidate documents, which then need to be re-ranked to obtain an ordered list based on relevance.

Recent efforts have explored the application of proprietary and open-source LLMs for ranking ~\citep{sun2023chatgpt,ma2023zero,pradeep2023rankvicuna,pradeep2023rankzephyr}, where the LLM is passed a combined list of passages directly as input and prompted to rank them based on a criteria.
Notably, only top-k candidates are passed as input to the LLM for re-ranking~\citep{zhang2023rank}.
In our work, we instruct an LLM to output an ordering of the different candidate papers (e.g.\ $[3]>[8]>[6]$) in descending order based on the relevance to the user-provided abstract.
Although existing re-ranking methods improve the ordering of candidate papers, they do not provide explicit justification for the relative rankings assigned. To improve the reliability of the system and offer a clear explanation for the model’s choices, our methodology also incorporates attribution capabilities, allowing us to identify specific textual elements contributing to relevance scores for different candidates.
Among works exploring the attribution capabilities in LLMs, \cite{yue2023automatic} focuses on automatically evaluating whether generated statements are fully supported by cited references.
\cite{cohenwang2024contextcite} present ContextCite, a method for attributing model generation to context, which can be applied on top of any existing language model to help verify generated statements, improve response quality, and detect poisoning attacks. Gradient-based techniques, such as Integrated Gradients~\citep{sundararajan2017axiomatic} and Saliency Maps~\citep{simonyan2014deep}, measure the contribution of each input token by computing gradients of the output with respect to input features; advances like SmoothGrad~\citep{smilkov2017smoothgrad} and DeepLIFT~\citep{shrikumar2017learning} improve these methods by reducing noise and enhancing accuracy.
Perturbation techniques involve modifying or occluding parts of the input and observing changes in the output to infer input significance~\citep{li2016understanding}, utilizing methods like Meaningful Perturbations~\citep{fong2017interpretable} and LIME~\citep{ribeiro2016should}. Despite advancements, challenges such as attribution leakage~\citep{adebayo2018sanity}, unreliability of saliency methods~\citep{kindermans2019reliability}, and complexities in attributing outputs in large models~\citep{ghorbani2019interpretation} persist.
Surveys like \cite{li2023surveylargelanguagemodels} discuss current methodologies and inherent challenges, while research by \cite{keeling2024attribution} examines the theoretical basis for attributing confidence to LLMs, raising concerns about the reliability of experimental assessment techniques.
In contrast to the discussed gradient-based attribution methods that are challenging to scale and perturbation-based approaches that need multiple passes through the model, we propose a straightforward prompting-based attribution approach that can be applied to any LLM agent, is readily scalable, and does not require multiple passes through the model.

\subsection{Literature Review Generation}
The concept of literature review generation using large language models (LLMs) is built upon the foundation laid by the Multi-XScience dataset proposed by \cite{lu-etal-2020-multi-xscience}. This dataset paves the way for the challenging task of multi-document summarization, specifically focusing on generating the related work section of a scientific paper. As underlined by \cite{lu-etal-2020-multi-xscience}, this approach favors abstractive models, which are well suited for the task. However, unlike the approach suggested by \cite{lu-etal-2020-multi-xscience}, our work introduces the use of intermediate plans to improve the quality of generated literature reviews. The empirical evidence presented in our study shows that our novel strategy outperforms the vanilla zero-shot generation previously championed by the Multi-XScience dataset~\citep{lu-etal-2020-multi-xscience}. (Note: This paragraph was entirely generated by GPT-4 following plan-based generation.\footnote{We use the plan: Please generate 5 sentences in 60 words. Cite @cite\_1 at line 1, 3 and 5. We postprocess to replace delexicalized tokens with latex commands. Outputs from other models are compared later in  Appendix (Tables \ref{table:qualitative} and \ref{table:qualitative-cite3}).})

Traditional methods for Natural Language Generation have typically employed a rule-based modular pipeline approach comprising of multiple stages of generation with intermediary steps of content planning (selecting content from input while also determining the structure of the output), sentence planning (planning the structure of sentences) and surface realization (surfacing the text in sentence) ~\citep{reiter1997building,stent-etal-2004-trainable,walker2007individual}. Our proposed plan-based prompting technique draws a parallel between the modern methods of end-to-end neural models for joint data-to-text generation with micro or content planning~\citep{gehrmann-etal-2018-end,puduppully-2019-data,puduppully-2021-macro} where we use plans to define the sentence structure of the generated output. While some recent works have explored planning in terms of devising actions~\citep{yang2022re3,song2023llm,wang2023plan}, prompting LLMs based on sentence plans have not been explored, to the best of our knowledge. We show two strategies of using plans 1.) The model generates the sentence plan as an intermediary step and conditions on this generated plan to output the final summary autoregressively. 2.) Humans can provide a ground-truth plan which results in an iterative setting, inherently providing controllability to the generated text where LLMs are susceptible to generating additional content.

Closely related to our work, \citet{gao2023enabling} generates answers for questions based on the citations from Wikipedia. Also related to our work, \cite{pilault-etal-2020-extractive} examined LLM-based abstractive summarization of scientific papers in the arxiv dataset of \cite{cohan-etal-2018-discourse}; however, their work was limited to creating the abstract of a single document. 
Perhaps the most similar prior prompting-based approach to our work is known as 0-shot chain-of-thought prompting~\citep{kojima2022large,zhou2022large} where a model is prompted with `Let's think step-by-step' (and similar prompts).

Additionally, Galactica has been developed to store, combine, and reason about scientific knowledge ~\citep{taylor2022galactica}. It outperforms existing models on various scientific tasks and sets new state-of-the-art results on downstream tasks. These findings highlight the potential of language models as a new interface for scientific research. However, the Galactica model was not developed to specifically address the problem of literature review assistance and it was not instruction fine-tuned to follow writing plans, and as such it suffered from the effect of hallucinating non-existent citations and results associated with imaginary prior work.\footnote{This sentence was inserted by the authors.} Recent works~\citep{rodriguez2024bigdocs, rodriguez2024starvectorgeneratingscalablevector, awadalla2024mint1tscalingopensourcemultimodal} have focused on building datasets multimodal of documents and scientific contents. However, our study focuses on exploring the zero-shot abilities of LLMs for literature review generation and proposes a novel strategy that includes generating an intermediate plan before generating the actual text. Our empirical study shows that these intermediate plans improve the quality of generated literature reviews compared to vanilla zero-shot generation. Furthermore, we ensure the validity of our experiments by using a new test corpus consisting of recent arXiv papers to avoid test set contamination. (Note: GPT-3.5 generated this paragraph with the 4th sentence added by the authors).

\section{Retrieval of Related Work}\label{sec:retrieval}

In this section, we discuss the creation of the corpus of arXiv papers to examine different retrieval strategies for finding related works for a given paper abstract using different academic and generic web search engines, including Semantic Scholar and Google Search.

\subsection{Dataset Construction}
We create two datasets that contain papers posted on arXiv in August and December 2023, respectively, starting with 1,000 papers from each month.
We use the arXiv wrapper in Python\footnote{\url{https://pypi.org/project/arxiv/}} to create RollingEval datasets.
We then filter out papers for which we were not able to retrieve 100 relevant paper results using LLM summarized keywords. We query the Semantic Scholar API available through the Semantic Scholar Open Data Platform \citep{lo-etal-2020-s2orc,kinney2023semantic} to search for the relevant papers. To get Google search results, we use SERP API,\footnote{\url{https://serpapi.com/}} specifically conditioned to leverage the ``site:arxiv.org'' parameter. This approach ensures the retrieval of search results are sourced solely from arXiv.org.

To combine results from multiple queries, we take the equal number of top results from each query to get a total of 100 papers. We took caution to avoid duplicate results from different queries. In case we are not able to retrieve a sufficient number of results from a query, we then take an equal number from the rest of the queries. This way we ensure that we always retrieve a candidate pool of 100 possible related work for each query paper. We pass these papers as queries to our literature review generation pipeline.
\begin{table}[b]
\centering
\resizebox{0.78\textwidth}{!}{
\begin{tabular}{lcc}
\toprule 
\textbf{Search type} & \textbf{RollingEval-Aug (\%)} & \textbf{RollingEval-Dec (\%)} \\
\midrule 
arXiv API (Single query) & 0.65 & 1.41 \\	
SERP API - Google Search (Single query) & 1.23  & 4.34\\	
Semantic Scholar API (Single query) & \textbf{3.93} & \textbf{4.76} \\	
\hdashline
arXiv API (Multiple queries) & 2.75 & 1.92 \\	
SERP API - Google Search (Multiple queries) & \textbf{6.80}  & 5.04\\	
Semantic Scholar API (Multiple queries) & 6.07 & \textbf{5.09} \\
\hdashline
SPECTER2 & 8.30 & 6.80 \\
Semantic Scholar API (Multiple queries) + SPECTER2 & \textbf{9.80} & \textbf{8.20} \\
\bottomrule 
\end{tabular}
}
\caption{We created two datasets to measure the efficacy of search using LLM-generated keywords - RollingEval-Aug and RollingEval-Dec.  We evaluate the \% of the ground truth references covered in the top 100 search list using LLM-based keyword search and different academic search engines. We note that using multiple queries gives us an edge over using a single query, and we obtain a similar coverage for Semantic Scholar and the SERP API-based Google Search.}
\label{table:results-search-api}
\end{table}
We now describe our two-step retrieval mechanism and provide its pseudo-code in Algorithm~\ref{alg:retrieval}.

\subsection{Retrieving Candidate Papers}
\label{subsec:retrieval}
\begin{algorithm}
\caption{Retrieval algorithm}
\label{alg:retrieval}
\begin{algorithmic}[1]
\REQUIRE Input abstract $a$
\STATE keywords = LLMKeywords($a$); // Generate keywords from the abstract using an LLM
\STATE candidate\_papers = SearchEngine(keywords); // Query a search engine to retrieve candidates
\STATE reranked\_papers = LLMRerank(candidate\_papers, $a$); // LLM-based reranking of candidates
\RETURN reranked\_papers
\end{algorithmic}
\end{algorithm}
To retrieve related work for a given paper abstract,
\textbf{first,} for each query abstract in the dataset, we prompt an LLM to generate keywords that we use as queries for a search API (refer to Figure \ref{fig:summarization-prompt} in the Appendix for the detailed prompt used for this task).
Importantly, we add a timestamp filter to the search API to retrieve papers published strictly before the publication date of the query paper.
In addition to evaluating multiple search engines, we also experiment with generating multiple queries\footnote{In our experiments, we generate three queries for each abstract.} from an LLM and various heuristics to combine the search results from each query (see Appendix \ref{appendix:implementation}).
We evaluate multiple general and academic search engines on the quality of the retrieved papers using coverage, which we define as the percentage of ground-truth papers retrieved by the search engine.

Table \ref{table:results-search-api} shows the coverage for different search engines and query heuristics. We note that using multiple queries achieves the highest coverage with comparable results for Semantic Scholar search and SERP API. However, at best, we retrieve just under 7\% of the ground truth papers. The low retrieval percentage of just under 7\% can be attributed to several factors. First, the task of finding related work for a given paper is inherently challenging due to the diverse styles and methods authors use in literature reviews. This stylistic variability means that a one-size-fits-all approach, such as generating search keywords and using a search engine, might not capture the nuanced criteria that a human expert would apply. Additionally, our retrieval process operates in a constrained setting, generating search keywords based solely on the paper's abstract. In theory, including more of the paper—such as the introduction or methodology—could provide richer context and lead to higher retrieval rates. However, this would not align with our intended use case: supporting researchers in the early stages of drafting when only limited, unpolished material is available.

We also explore an embedding-based strategy for retrieval using SPECTER embeddings~\citep{cohan2020specter}.
SPECTER uses a contrastive learning approach to train a SciBERT \citep{beltagy-etal-2019-scibert} model based on a triplet loss to discriminate between related versus unrelated papers. SciBERT is a BERT-like transformer model~\citep{devlin2018bert} trained on scientific text. More recently, \cite{singh2022scirepeval} further extended the contrastive learning approach of SPECTER to better deal with multiple tasks of relevance to scientific papers, including citation prediction, leading to SPECTER2. Since SPECTER2 provides a large knowledge base of 150M document embeddings of scientific articles, we explore embedding-based retrieval of candidate papers, where we obtain the top-$k$ papers based on the cosine similarity between the query abstract and the corpus of candidate papers in the SPECTER2 dataset.


\subsection{Re-ranking Candidate Papers}
\textbf{Next,} given a list of retrieved papers, we explore re-ranking the list using an LLM.
The retrieved abstracts and the original query abstract are used as input to an LLM Re-ranker, which provides a listwise ranking of the candidate papers based on the relevance to the query abstract. 
We explore different strategies for reranking the candidates, detailed as follows:
\textbf{a) {Instructional permutation generation}}: we use the approach by~\citet{sun2023chatgpt}, which prompts the model to directly generate a permutation of the different candidate papers, thus producing an ordered list of preferences against providing intermediate scores; \textbf{b) SPECTER2 embeddings}: We use SPECTER2 embeddings as an alternative to prompting-based strategies for reranking, where we rank the candidate papers based on their cosine distances to the SPECTER2 embedding of the query abstract (see Appendix~\ref{appendix:implementation} for more details on SPECTER2 implementation); and \textbf{c) Debate Ranking with Attribution (Ours)}: our prompting-based approach that builds on the work of~\citet{rahaman2024language}, where we pass each candidate paper's abstract along with the query abstract and instruct the LLM to (1) generate arguments for and against including the candidate paper and (2) output a final probability of including the candidate based on the arguments. Crucially, we add an attribution step to this ranking module, where we instruct the LLM to extract verbatim sentences from the candidate abstract that support the arguments, and we re-prompt the LLM if the extracted sentences are not present in the candidate abstract. 

\subsection{Retrieval and Re-ranking Experiments}

We use an ensemble of search engines to retrieve candidates based on an abstract. We now describe the search engines used in our approach.
Based on Table~\ref{table:results-search-api}, we select the Semantic Scholar (S2) API as the search engine to retrieve search results using LLM-generated keywords. While the SERP API provides access to a broader set of search results, it is expensive, making it less practical for large-scale retrieval. Additionally, the S2 API is specifically designed for academic literature, offering strong performance comparable to the SERP API.
As discussed in Section~\ref{subsec:retrieval}, we also explore SPECTER2 embeddings for retrieval and compare it with the different prompting-based strategies.

\begin{figure}[t]
  \centering
  \begin{subfigure}{0.8\textwidth}
      \includegraphics[width=\textwidth]{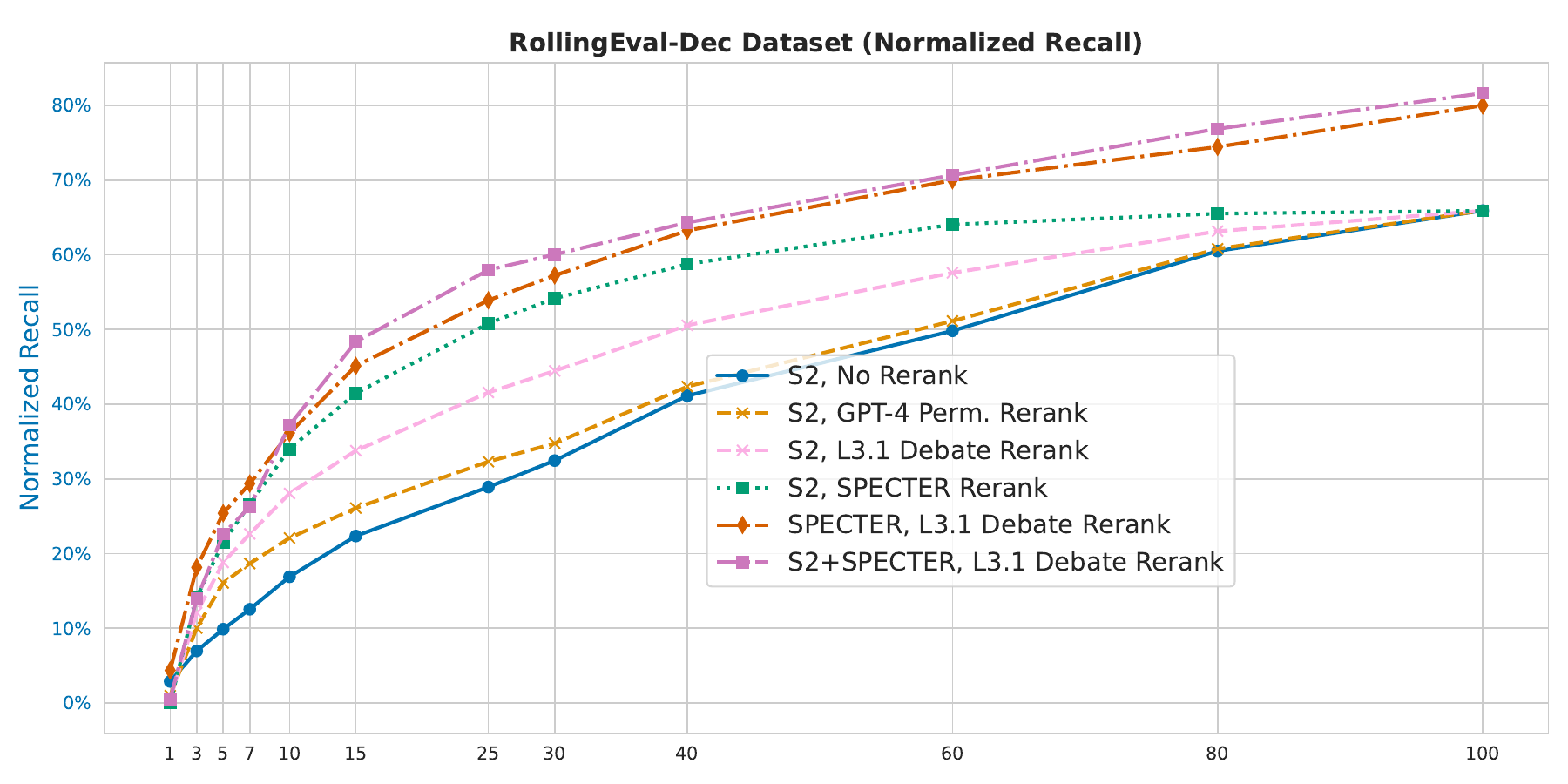}
  \end{subfigure}
  \begin{subfigure}{0.8\textwidth}
      \includegraphics[width=\textwidth]{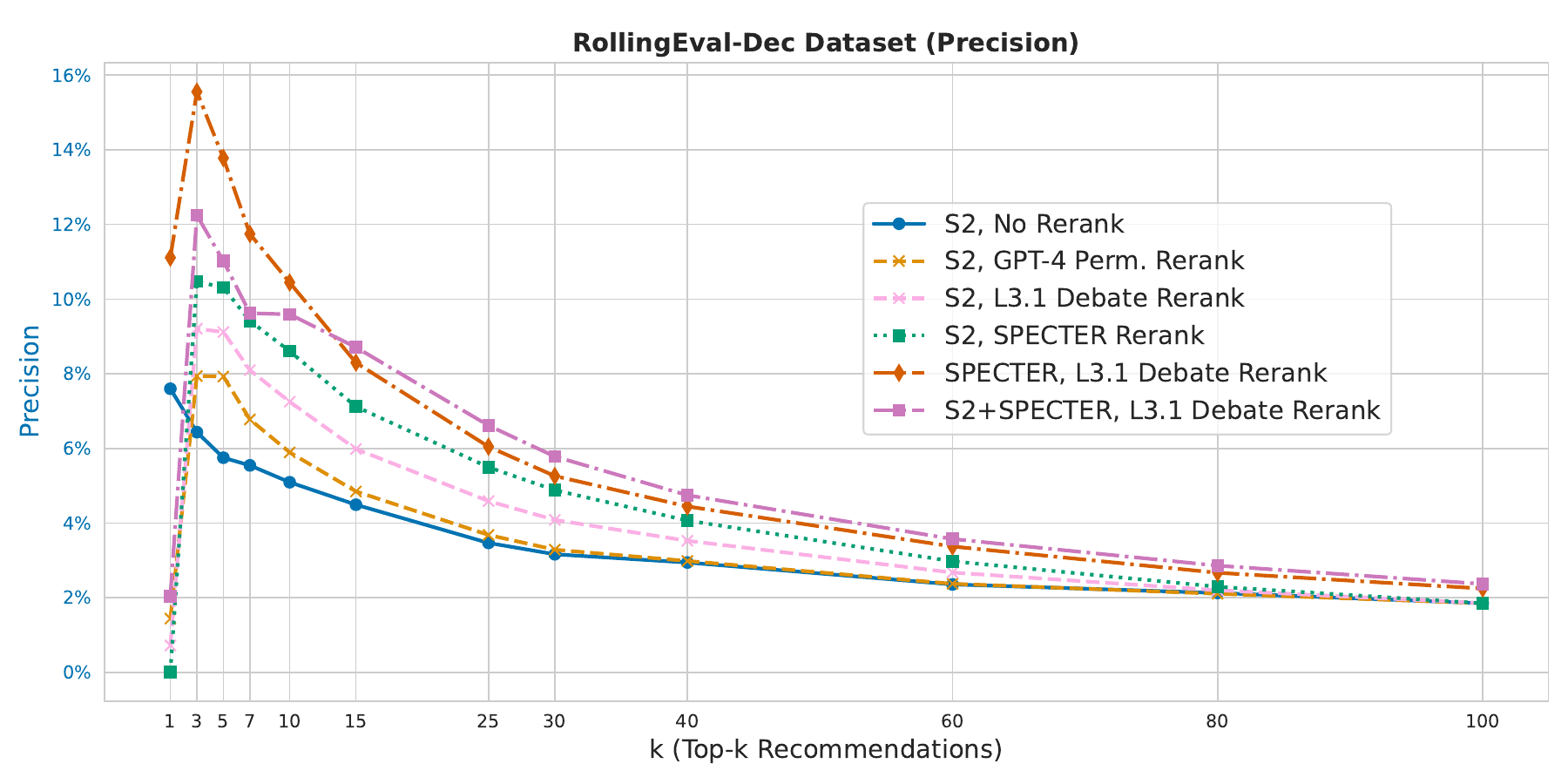}
  \end{subfigure}
     \caption{Effect of re-ranking strategies on the RollingEval-Dec dataset. We use the entire dataset ($n=500$) and set $k=100$ for these experiments. We evaluate the Precision and Normalized Recall of the re-ranked results with embedding-based ranker (SPECTER2) outperforming GPT-4 based re-ranking. We find a similar pattern for the RollingEval-Aug dataset, as shown in Appendix (Figure \ref{appendix-figure:retrieval-pr-curves}). \textbf{Note:} The first part in the legend denotes the search database for retrieval, and the second denotes the re-ranking mechanism.}
     \label{figure:retrieval-pr-curves}
\end{figure}

We experiment with different combinations of search engine and the retriever method, and present our results in Figure \ref{figure:retrieval-pr-curves}.
We present precision and normalized recall at different values of top-$k$ recommendations, where we calculate normalized recall as the proportion of the number of ground truth papers retrieved (instead of all ground truth papers).
Formally, normalized recall and precision follow the definitions:

\begin{equation}
\text{Normalized Recall@k} = \frac{ |\text{Retrieved@k} \cap \text{Ground Truth}|}{|\text{Retrieved} \cap \text{Ground Truth}|}; \text{Precision@k} = \frac{|\text{Retrieved@k} \cap \text{Ground Truth}|}{k}
\end{equation}

where ``Retrieved'' denotes the set of \textit{all} candidate papers, ``Retrieved@k'' denotes the set of top $k$ candidate papers, and ``Ground Truth'' denotes the set of papers cited by the query paper.
Unlike standard recall, which measures how many ground truth citations are retrieved, Normalized Recall@k measures how effectively the retrieval method prioritizes the most relevant papers in the top-$k$. By normalizing over the total relevant papers retrieved at any rank, this metric helps evaluate ranking quality independently of retrieval coverage.
We include a working example in Appendix~\ref{app:recall-example} to further demonstrate the difference between normalized recall and standard recall.

From Figure~\ref{figure:retrieval-pr-curves}, we note that debate ranking significantly outperforms permutation ranking, as denoted by the higher precision and normalized recall at smaller values of top-k recommendations; however, SPECTER embeddings outperform both prompting-based strategies.
Next, the higher precision and normalized recall values for SPECTER and S2+SPECTER settings suggest that SPECTER is also an excellent search engine.

In Table \ref{table:results-search}, we examine the behaviour of the GPT-4 reranking approach in more detail. Using GPT-4 as a reranker in the manner discussed above produces an incomplete list 41\% of the time. In 3\% cases, GPT-4 produces a list with repeated values and, in some rare cases, with garbage values or numbers (e.g.\ 2020). We conclude that this strategy of using GPT-4 for reranking is brittle.

\subsection{Positive Effect of Attribution Verification}

\begin{figure}
  \centering
    \includegraphics[width=0.7\textwidth]{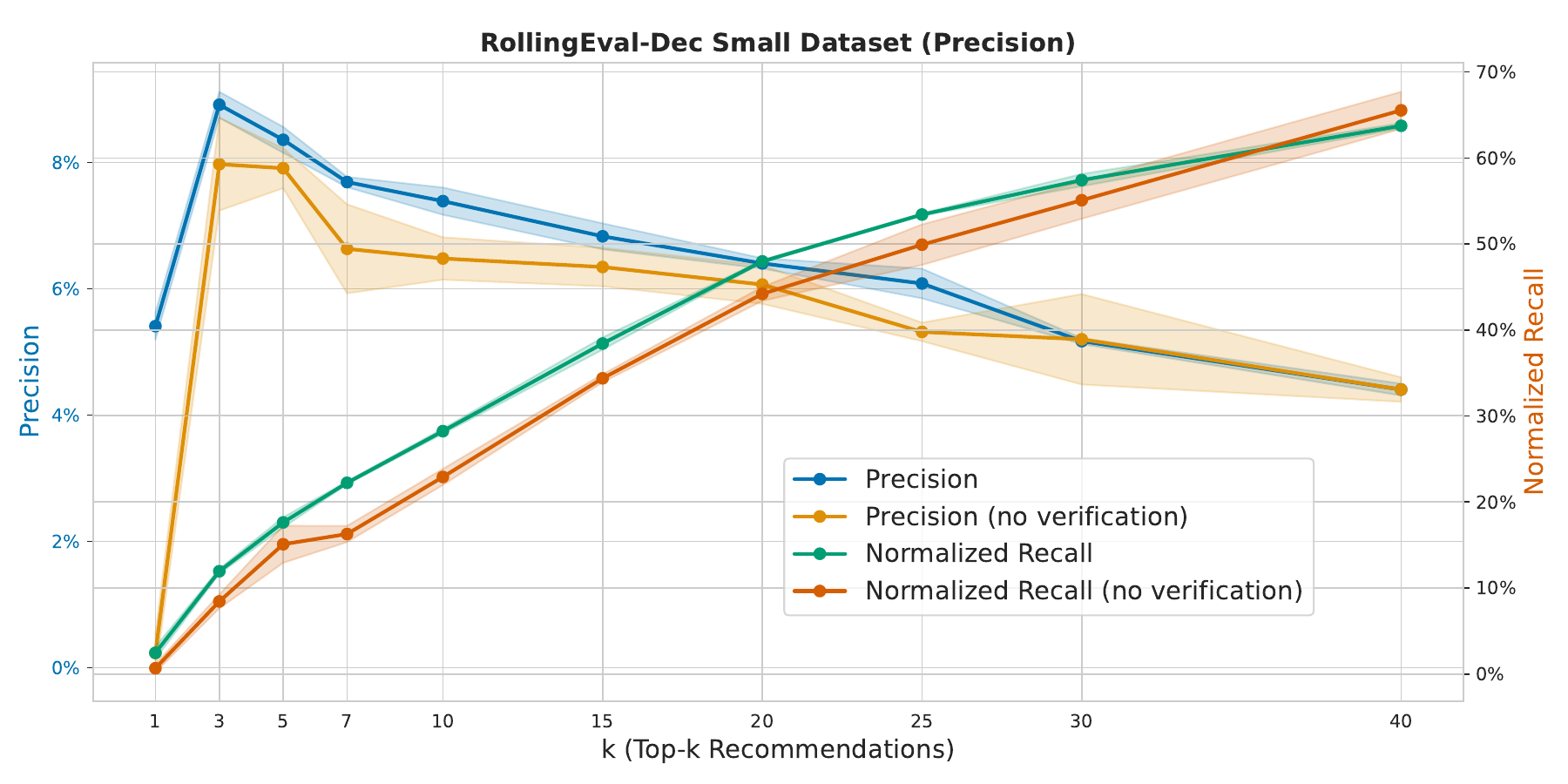}
    \caption{The effect of removing the referenced content verification step in our debate ranking strategy. We plot precision and normalized recall for two variants of the debate ranking strategy. For this ablation study, we select a smaller subset of $n=100$ query abstracts, set $k=40$, and repeat the experiment for three random seeds. We plot the mean and show the standard deviation as the shaded region. We find that the precision and normalized recall drop slightly upon removing the verification step. This difference is significant (as determined by the t-test,) indicating that the verification step is crucial for the success of the debate ranking strategy.}
    \label{figure:ablation-pr-curves}
\end{figure}

We conduct an ablation study on the first 100 papers of the larger set of 500, and focus on the top $k=40$ papers, which is representative of the typical number of papers cited in the Machine Learning community. 
In Figure \ref{figure:ablation-pr-curves} we show the result of removing the verification step in our debate re-ranking strategy, i.e.\ in this ablation, we do not check if the sentences extracted by the LLM are indeed present in the candidate paper abstract. 
We find that removing this kind of attribution verification leads to a drop in the precision and normalized recalls, especially for lower $k$ values.
We perform a t-test to test the significance of the drop and find that for both precision and normalized recall, the drop is significant, with $p$-values of $\num{4.7e-4}$ and $\num{1.9e-6}$ for precision and normalized recall curves, respectively.
This indicates that proper attribution also allows the LLM to provide a more accurate ranking of candidates.

\begin{table}[ht]
\centering
\resizebox{0.6\textwidth}{!}{
\begin{tabular}{lS[table-format=2.1,table-number-alignment = center]S[table-format=2.1,table-number-alignment = center]}
\toprule 
\textbf{Ranker Predictions} & \textbf{RollingEval-Aug (\%)} & \textbf{RollingEval-Dec (\%)} \\
\midrule 
Complete Ranked list & 55.1 & 59.7\\
Incomplete list & 41.5 & 40.2\\	
Repeated Value & 3.3 & 0.1\\	
\bottomrule 
\end{tabular}
}
\caption{Error modes of GPT-4 based ranking. When using GPT-4 to provide ranks, it suffers from multiple issues. While recent works like~\cite{sun2023chatgpt} explore LLMs like GPT-4 for 0-shot rank predictions, we find a tendency of GPT-4 to produce an incomplete list or repeated values in the re-ranked order list with some rare cases of garbage values.}
\label{table:results-search}
\end{table}

\section{Literature Review Generation}

\textbf{Plan Based Generation Approach \& Model Variants.} \label{sec:approach}
We now focus on generating the related work section of a scientific document from a user-supplied list of papers. In a real-world scenario, this list might be obtained through traditional means, from the above-mentioned automated methods, or some combination.
%
We evaluate several dimensions of writing quality in the generated text. Importantly, while modern LLMs can yield seemingly well-written text passages, ``hallucinations'' remain a problem and can be particularly egregious when LLMs are used in scientific writing~\citep{Athaluri2023ExploringTB}. The hallucination of statements not entailed by the contents of cited papers and the hallucinations of imaginary papers that do not exist is a well-known issue of even the most powerful LLMs. We use ideas from retrieval augmented generation (RAG) techniques~\citep{lewis2020retrieval} and instruction prompting to address the key problem of hallucinations.
Our work also aims to increase the number of papers from the desired set that are indeed discussed (the coverage). 

\begin{figure*}[ht]
\centering
\includegraphics[width=0.7\linewidth]{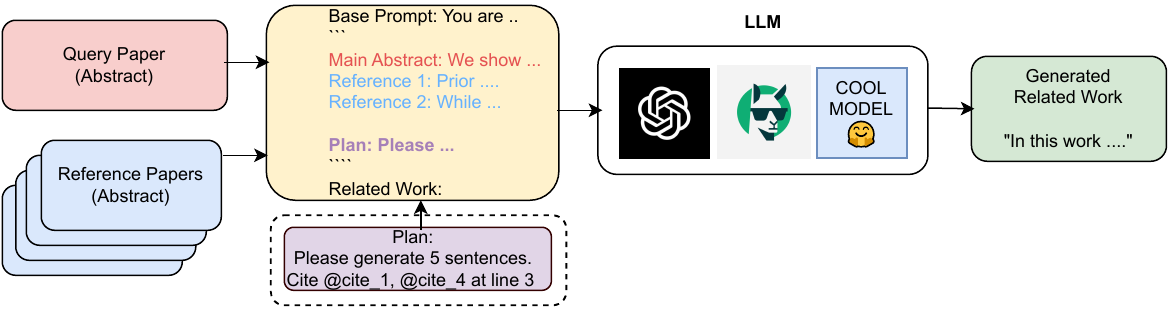}
\caption{Pipeline of generation task where the model needs to generate the related work of the query paper conditioned on reference papers. Our method employs an optional plan --- shown by the dotted purple box, either generated by the model or appended to the prompt.}
\label{fig:pull-figure}
\end{figure*}

We present our general framework and problem setup in Figure~\ref{fig:pull-figure}. We use the abstract of a query paper --- the one for which we generate a literature review, along with the abstracts of the set of papers to be cited (the retrieved abstracts of reference papers) to generate the related work section of the query paper. For evaluation, our approach relies on prompting LLMs in different ways and measuring the similarity of the generated literature review text to ground truth literature reviews found within a corpus of recent scientific papers --- i.e.\ ones not used in the training set of the underlying LLMs. We use both automated methods and human evaluations in our experiments below.

We propose to further decompose the writing task to increase passage quality, factual accuracy, and coverage. We examine different strategies for generating a \emph{writing plan}, a line-by-line description including citations of the passage to write. These writing plans also give authors (users) control over the output passages. This is likely essential in practice to meet author preferences and possible publication constraints. These plans are defined and generated in such a way that the user can interact with and edit them if desired, or they can be used as an automatically generated intermediate (but human-understandable) representation. We now describe our proposed methods, their use in practice and their evaluation in more detail. 

In what we refer to as plan-based generation (\textbf{Plan}), a model is prompted with a known (user-provided) plan to produce X sentences in Y words and cite references on specific lines.
These plans are obtained from the ground truth data and serve as a proxy for a user who desires text to be written with specific constraints. This kind of Plan strongly guides generated text to the user's desires (as observed by the final ground truth text). During evaluation, this might be considered a form of teacher-forcing at the structural level. An example of the format of these plans is provided below:

\vspace{2mm}
\noindent\fbox{%
    \parbox{0.99\linewidth}{%
\scriptsize{\texttt{{Please generate \{num\_sentences\} sentences in \{num\_words\} words. Cite \{cite\_x\} at line \{line\_x\}. Cite \{cite\_y\} at line \{line\_y\}.
        }}}
    }%
}\\

\textbf{Prompted plan.} 
The previous method replicates the scenario of a detailed user-provided plan. However, one can also generate such plans automatically from an LLM. The model is prompted first to generate a plan of sentences and citations, which it would then condition upon to generate the final related work text. When used as an interactive tool, we envision the user might start with a suggested plan, see the corresponding generated full literature review text, and then iteratively edit the plan and regenerate the result. See our Appendix (Figures \ref{fig:prompt-0-shot}, \ref{fig:prompt-plan} and \ref{fig:prompt-learned-plan}) for the differences in the prompts used in our experiments. 
We also experiment with two other strategies in which researchers could prompt the model: 

\textbf{Per cite.} We first use a two-stage strategy to generate content relevant to each cited paper. In the first stage, the model is prompted to generate related works in 1-2 lines for each individual reference citation. All the outputs for different citations are combined together to form the generated related work. In the second stage, the LLM summarizes and paraphrases the output of the first stage. 

\textbf{Sentence by sentence} Based on the Ground truth (GT) related work and the citation on each line, we prompt the model to generate one sentence conditioned on the abstract, the reference cited in that line, and the generated draft so far. In the absence of a citation or at the start, the model is prompted only with the abstract and draft of the generated work till now.

\section{Generation Experiments.}
\label{sec:experiments}

For the following studies on generating related work, we introduce an additional corpus. We extend the Multi-XScience corpus~\citep{lu-etal-2020-multi-xscience} to include the full text of research papers. We also reuse the RollingEval-Aug introduced in Section~\ref{sec:retrieval}.
We use HuggingFace Transformers~\citep{wolf2019huggingface} and PyTorch
~\citep{paszke2017automatic} for our experiments\footnote{Code can be accessed at \url{https://github.com/LitLLM/litllms-for-literature-review-tmlr}} and calculate ROUGE scores~\citep{lin-2004-rouge} using the Huggingface's \texttt{evaluate} library. Details on the dataset and the implementation are in Appendix~\ref{appendix:datasets} and~\ref{appendix:implementation}, respectively. Similar to~\cite{lu-etal-2020-multi-xscience}, we extract the ground truth cited references as the relevant papers and evaluate only the generated outputs from different systems. Since ROUGE score only measures token-level similarity and does not account for semantic meaning, we also report BERTScore and Llama-3-Eval in Table~\ref{table:results-rw-2308}.
We use BERTScore, an embedding-based evaluation metric, to account for the semantic meaning of two texts during evaluation as well.
On the other hand, we use Llama-3-Eval, an open-source variant of the widely used G-Eval metric~\citep{liu-etal-2023-g}, as G-Eval has been shown to correlate better with human preferences.


\subsection{Generation Baselines}

\textbf{Extractive baselines} As in~\cite{lu-etal-2020-multi-xscience}, we report the performance of LexRank~\citep{erkan2004lexrank} and TextRank~\citep{mihalcea2004textrank}. We also create a simple one-line extractive baseline which extracts the first line of the abstract and combines all the citations to form the output.

\textbf{Abstractive finetuned baselines} We use the model outputs of Hiersum~\citep{liu2019hiersumm} and Pointer-Generator~\citep{see2017pointer} from~\cite{lu-etal-2020-multi-xscience} for abstractive finetuned baselines. We also reproduce the finetuned PRIMER~\citep{xiao2021primera} model (considered to be the SOTA).

\textbf{Abstractive zero-shot baselines} We use the zero-shot single-document abstractive summarizers FlanT5~\citep{chung2022scaling} and LongT5~\citep{guo-etal-2022-longt5} based on the T5 architecture~\citep{raffel2020exploring}. Since Galactica~\citep{taylor2022galactica} is trained on documents from a similar domain, we include it 
along with Falcon-180B~\citep{almazrouei2023falcon}.

\textbf{Open and closed source models}
We use different chat versions (7B, 13B, 70B) of Llama 2-Chat\footnote{We refer to Llama 2-Chat models as Llama 2 and GPT-3.5-turbo as GPT-3.5 for brevity.}~\citep{touvron2023llama2} as zero-shot open-source LLM baselines. For closed-source models, we
evaluate zero-shot both GPT-3.5-turbo~\citep{brown2020gpt3} and GPT-4~\citep{openai2023gpt}. Since they perform best in our initial evaluations, we use the closed-source models in combination with the different generation strategies (Per-cite, Sentence by sentence, plan-based, and learned plan) from Section~\ref{sec:approach}.

\begin{table*}[ht]
\centering
\resizebox{0.75\textwidth}{!}{
\begin{tabular}{llccc}
\toprule
\textbf{Model Class} & \textbf{Model} & \textbf{ROUGE1 $\uparrow$} & \textbf{ROUGE2 $\uparrow$} & \textbf{ROUGEL $\uparrow$} \\
\midrule 
\multirow{3}{*}{Extractive}
& One line baseline & 26.869 & 4.469 & 14.386 \\
& LexRank & 30.916 & 5.966 & 15.916 \\
& TextRank & 31.439 & 5.817 & 16.398 \\
\midrule 
\multirow{3}{2cm}{Abstractive Finetuned} 
& Hiersum  & 29.861 & 5.029 & 16.429 \\
& Pointer-Generator  & 33.947 & 6.754 & 18.203 \\
& PRIMER  & 26.926 & 5.024 & 14.131 \\
\midrule 
\multirow{4}{2cm}{Abstractive 0-shot} 
& Long T5  & 19.515 & 3.361 & 12.201 \\	
& Flan T5  & 21.959 & 3.992 & 12.778 \\
& Galactica-1.3B & 18.461 & 4.562 & 9.894 \\
& Falcon-180B & 22.876 & 2.818 & 12.087 \\
\midrule 
\multirow{3}{2cm}{Open-source 0-shot} 
& Llama 2-Chat 7B (No plan) & 24.636 & 5.189 & 13.133 \\
& Llama 2-Chat 13B (No plan) & 26.719 & 5.958 & 13.635 \\
& Llama 2-Chat 70B (No plan) & 28.866 & 6.919 & 14.407 \\
& LLama-3.1-70B (No Plan) & 33.289 & 8.050 & 15.898 \\
\midrule 
\multirow{3}{3cm}{Closed-source 2-stage} 
& GPT-3.5-turbo (Per cite) 1st stage & 26.483 & 6.311 & 13.718 \\	
& GPT-3.5-turbo (Per cite) 2nd stage & 24.359 & 5.594 & 12.859 \\
& GPT-3.5-turbo (Sentence by sentence) & 31.654 & 6.442 & 15.577 \\
\midrule 
\multirow{2}{2.5cm}{Closed-source 0-shot}
& GPT-3.5-turbo (No plan) & 29.696 & 7.325 & 14.562 \\
& GPT-4 (No plan) & 33.213 & 7.609 & 15.798 \\
\midrule 
\multirow{4}{2cm}{Plan}
& Llama 2-Chat 70B (Prompted plan) & 30.389 & 7.221 & 14.911 \\	
& GPT-3.5-turbo (Prompted plan) & 32.187 & 7.788 & 15.398 \\
& GPT-4 (Prompted plan) & 34.819 & 7.892 & 16.634 \\
& Llama 2-Chat 70B (Plan) & 34.654 & 8.371 & 17.089 \\
& GPT-3.5-turbo (Plan) & 35.042 & 8.423 & 17.136 \\
& GPT-4 (Plan) & 37.198 & 8.859 & 18.772 \\
& Llama-3.1-70B (Plan) & 35.575 & 9.406 & 18.772 \\
\bottomrule
\end{tabular} }
\caption{Zero-shot results for different models on the Multi-XScience dataset.}
\label{table:results-0shot}
\end{table*}

\section{Results and Observations}
\label{section:observations}

From Table~\ref{table:results-0shot}, we first note that unsupervised extractive models provide a strong baseline compared to abstractive 0-shot single document summarization baselines. Fine-tuning these abstractive models on Multi-XScience (released initially with the benchmark) improves performance at least to the level of extractive models. We reproduce the PRIMER model using their open-source code but find lower-than-reported results. As such, we consider the Pointer-Generator method to be the current state-of-the-art (SOTA).

\begin{table*}[ht]
\centering
\resizebox{0.8\textwidth}{!}{
\begin{tabular}{lccccc}
\toprule
\textbf{Model} & \textbf{ROUGE1 $\uparrow$} & \textbf{ROUGE2 $\uparrow$} & \textbf{ROUGEL$\uparrow$} & \textbf{BERTScore$\uparrow$} & \textbf{Llama-3-Eval$\uparrow$} \\
\midrule 
CodeLlama 34B-Instruct & 22.608 & 5.032 & 12.553 & 82.418 & 66.898 \\
CodeLlama 34B-Instruct (Plan) & 27.369 & 5.829 & 14.705 & 83.386 & 67.362 \\
\hdashline
Llama 2-Chat 7B & 23.276 & 5.104 & 12.583 & 82.841 & 68.689 \\
Llama 2-Chat 13B & 23.998 & 5.472 & 12.923 & 82.855 & 69.237 \\
Llama 2-Chat 70B & 23.769 & 5.619 & 12.745 & 82.943 & 70.980 \\
\hdashline
GPT-3.5-turbo (0-shot) & 25.112 & 6.118 & 13.171 & 83.352 & 72.434 \\
GPT-4 (0-shot) & 29.289 & 6.479 & 15.048 & 84.208 & 72.951 \\
\hdashline
Llama 2-Chat 70B (Plan) & 30.919 & 7.079 & 15.991 & 84.392 & 71.354 \\
GPT-3.5-turbo (Plan) & 30.192 & 7.028 & 15.551 & 84.203 & 72.729 \\
GPT-4 (Plan) & 33.044 & 7.352 & 17.624 & 85.151 & 75.240 \\
\bottomrule
\end{tabular}
}
\caption{Zero-shot results on the proposed RollingEval-Aug dataset.}
\label{table:results-rw-2308}
\end{table*}

\begin{table}[ht]
\centering
\resizebox{0.64\textwidth}{!}{
\begin{tabular}{lS[table-format=2.2]S[table-format=1.2]S[table-format=2.]S[table-format=2.1]S[table-format=1.2]S[table-format=2.]}
\toprule
\multicolumn{1}{l}{\multirow{2}{*}{\textbf{Model}}} & \multicolumn{3}{c}{\textbf{Multi-XScience}} & \multicolumn{3}{c}{\textbf{RollingEval-Aug}} \\ \cline{2-7} 
\multicolumn{1}{l}{}                       & \multicolumn{1}{c}{\textbf{\% $\uparrow$}} & \multicolumn{1}{c}{\textbf{Mean $\downarrow$}} & \multicolumn{1}{c}{\textbf{Max $\downarrow$}} & \multicolumn{1}{c}{\textbf{\% $\uparrow$}} & \multicolumn{1}{c}{\textbf{Mean $\downarrow$}} & \multicolumn{1}{c}{\textbf{Max $\downarrow$}} \\ 
\midrule 
GPT-3.5-turbo (Plan) & 4.73 & 3.65 & 17 & 3 & 4.7 & 16 \\
Llama 2-Chat 70B (Plan) & 19.04 & 2.66 & 22 & 17.4 & 2.72 & 18 \\
GPT-4 (Plan) & 60.7 & -0.01 & 8 & 70.6 & 0.16 & 5 \\
\bottomrule
\end{tabular}
}
\caption{We show \% of responses with the same number of lines as the plan for both datasets. Here we also show the mean and max difference in lines generated by the model vs.\ the original plan. -ive implies that a model generated fewer lines than the plan. We find GPT-4 to follow the plan more closely compared to Llama 2 and GPT-3.5.}
\label{table:results-plan-diff}
\end{table}

Single-document summarizers (LongT5, Flan T5) perform poorly in the zero-shot settings with limited ability to cite references. We are limited in the prompt we can provide (because of the training prompts) and resort to ``Summarize the text and cite sources.'' Galactica's performance is encouraging compared to other models in the same group, but inspecting its output reveals that it generates the whole introduction of the paper instead of the related work. The model is very sensitive to the prompts used (mostly as suffixes) and struggles to follow instructions. Falcon 180-B, on the other hand, tends to hallucinate user turns and considers this task as multiple turns of user-system exchange, even though we prompted to generate relevant outputs.

All recent versions (7B, 13B, 70B) of zero-shot Llama 2 models underperform the supervised Pointer-Generator baseline (except for 70B on ROUGE2) and their GPT counterparts. All Llama 2 models tend to produce output in bullet points and also provide references. 
We find that closed-sourced models like GPT-3.5-turbo and GPT-4 achieve SOTA in the zero-shot setting. However, the proposed sentence-by-sentence and per-citation strategies deteriorate the performance of GPT models which tends to cover all related concepts hierarchically.\footnote{We validated these strategies only on GPT-3.5 due to the high incurred cost with GPT-4 models.} 

Our teacher-forced plan-based framework improves the scores over the 0-shot baseline for both closed-sourced (GPT-3.5 and GPT-4) and open-sourced LLMs, with Llama 2 70B achieving similar scores as GPT-3.5 on both the original Multi-XScience and the new RollingEval-Aug dataset (in Table~\ref{table:results-rw-2308}). In Table~\ref{table:results-rw-2308}, we also notice that BERTScore and Llama-3-Eval exhibit the same trends as ROUGE scores except in the case where GPT-3.5-turbo (Plan) obtains a higher Llama-3-Eval score than Llama 2-Chat 70B (Plan).
These values also showcase the weaker discerning power of BERTScore compared to Llama-3-Eval as all the models achieve a high BERTScore between 82-85\%.
Llama 2 70B gets more uplift with the plan compared to GPT models where manual inspection reveals fewer hallucinations in the outputs (see qualitative results in Table \ref{table:qualitative} in Appendix using our abstract). In Table \ref{table:results-plan-diff}, we evaluate the controllability of LLM-based generation using sentence plans and find that GPT-4 tends to follow the plan more closely. It follows the exact plan 60\% of the time, often producing fewer sentences than provided. Llama 2 70B comes in second place in following the plan instructions and GPT-3.5 struggles to follow the plan precisely. We also experiment with a learned plan strategy where the model first generates a plan and then autoregressively generates the output. Though it improves the results over 0-shot baseline, it does not outperform the teacher-forced plan generation in terms of automatic metrics. 
There is a considerable drop in performance on RollingEval-Aug dataset compared to the original Multi-XScience in terms of ROUGE1/2. It gives more credibility to the hypothesis that the Multi-XScience test set is in the training data of these LLMs and/or that there is a shift in the distribution of these more recent papers. Nevertheless, we find similar scores and ranking patterns between models as for Multi-XScience. We provide other experiments related to fine-tuning, longer context, and CodeLLM in Appendix \ref{appendix:experiments}, and we provide cost estimates for different methods in Appendix~\ref{appendix:costs}.

\textbf{Coverage and human evaluation:} We evaluate coverage as the percentage of model outputs with the same number of citations as ground truth (identified using regex on the delexicalized citation in the generated related work). Table~\ref{table:results-llm-eval} shows the efficacy of plan-based approaches. All plan models provide more coverage than their counterparts, with GPT-4 achieving 98\% in covering all the citations. The largest uplift is for (vanilla) 0-shot Llama 2 70B. Using a plan raises its coverage from  59\% to 82\%. Similar to results in Table~\ref{table:results-plan-diff}, we find that the coverage of GPT-3.5 does not improve much. 

\begin{table}[ht]
\centering
\resizebox{0.5\textwidth}{!}{
\begin{tabular}{lcS[table-format=2.1,table-number-alignment = center]}
\toprule 
\textbf{Model} & \textbf{Coverage $\uparrow$} & \textbf{Avg.\ words} \\
\midrule 
Llama 2-Chat 70B (0-shot) & 59.31\% & 284.65\\	
Llama 2-Chat 70B (Plan) & 82.62\% & 191.45\\	
GPT-3.5-turbo (0-shot) & 63.11\% & 293.69\\	
GPT-3.5-turbo (Plan) & 68.03\% & 202.81\\	
GPT-4 (0-shot) & 91.34\% & 215.15\\	
GPT-4 (Plan) & 98.52\% & 125.10\\	
\bottomrule 
\end{tabular}
}
\caption{Coverage (in \%) on the Multi-XScience dataset defines the number of citations covered in the generated response.}
\label{table:results-llm-eval}
\end{table}

\begin{figure}
     \centering
     \begin{subfigure}[b]{0.41\textwidth}
         \centering
         \includegraphics[width=\textwidth]{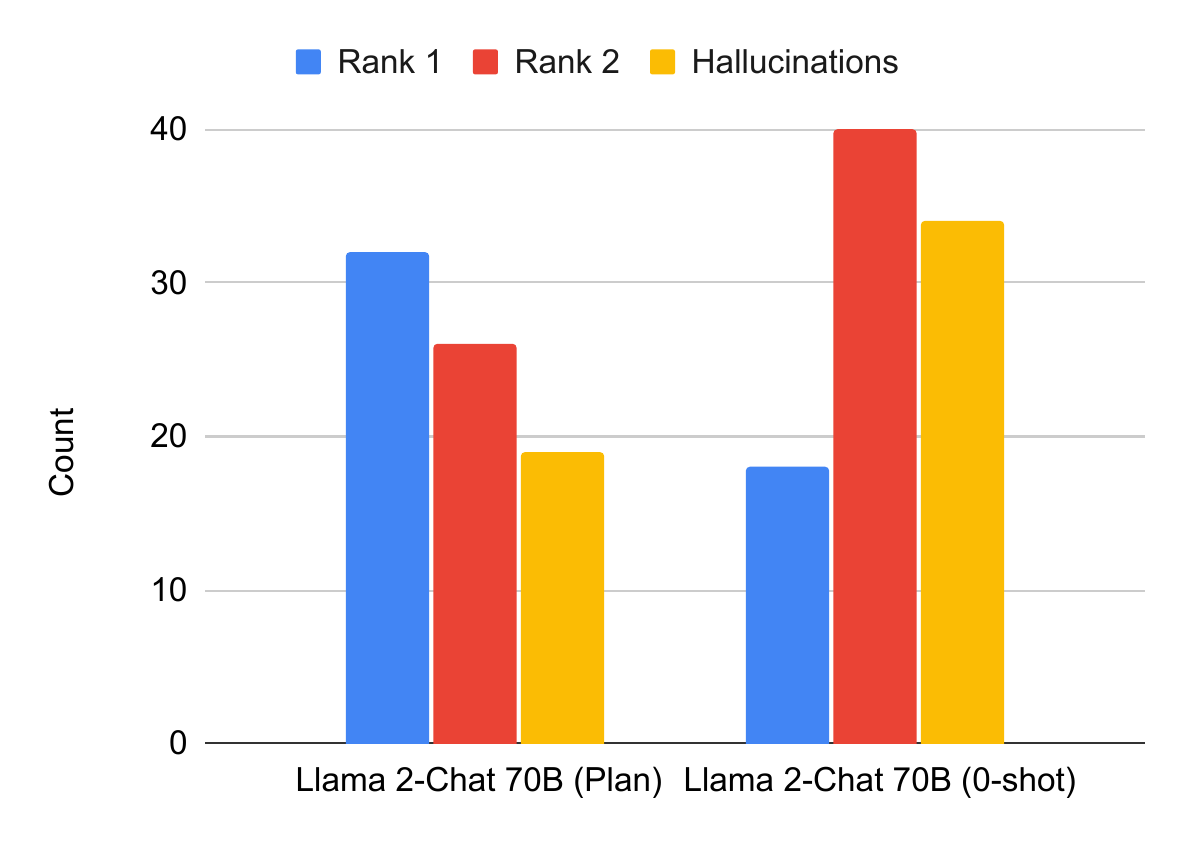}
     \end{subfigure}
     \hfill
     \begin{subfigure}[b]{0.41\textwidth}
         \centering
         \includegraphics[width=\textwidth]{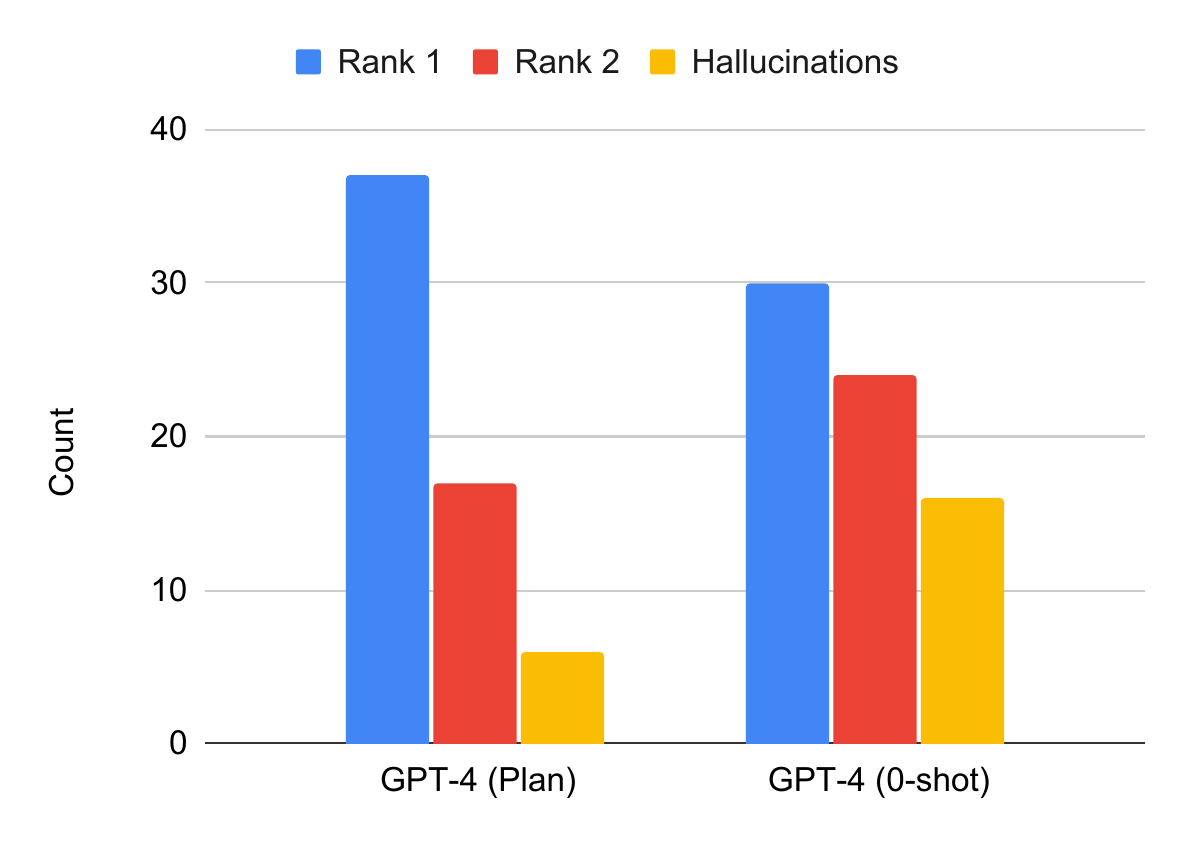}
     \end{subfigure}
        \caption{Human evaluation study where annotators ranked the generations of 0-shot models with their sentence-plan-based counterparts. On the Y-axis, we show counts from an overall sample size of 58 annotations for Llama 2-Chat and 54 for GPT-4 (where ranking ties are allowed). We see a reduction of 58.6\% cases of hallucinations to 32.7\% for Llama 2-Chat and 29.6\% to 11.6\% for GPT-4 using plan-based prompting.}
        \label{figure:results-human-eval}
\end{figure}


We also run a human evaluation study using 6 expert annotators. They rank model generated outputs for 160 papers with 3 citations\footnote{This was done to reduce cognitive load on annotators to read abstracts of 4+ papers and 2 model outputs. Annotators have research experience in machine learning at the Masters or Doctorate level.} where we show the abstract of the query paper, cited references, and the model outputs of 0-shot vs plan counterparts side-by-side. We randomly selected 80 examples each for GPT-4 and Llama 2-Chat comparisons. Experts had the flexibility to choose rank 1 for one or both the models, otherwise they could select rank 2 for each. We also ask the annotators to identify hallucinations, i.e.\ which model generated content not from the abstracts. The interface is described further and shown in the Appendix (Figure \ref{fig:interface-human-eval}). Out of 160, we find agreement among the annotators for 112 examples (54 for GPT4 and 58 for Llama 2-Chat). The results in Figure \ref{figure:results-human-eval} show that humans rank generated response to be significantly better\footnote{We used McNemar test~\citep{lachenbruch2014mcnemar} to measure statistical significance.} for Llama 2-Chat Plan based model where it was ranked at the top 32 times compared to 18 times for 0-shot. We can also observe a similar trend for GPT-4, where annotators rank GPT-4 Plan-based output 37 times at Rank 1 compared to 30 for GPT-4 0-shot. In terms of hallucinations, we find significant reductions in the hallucination for plan-based Llama 2-Chat models from 34 to 19 instances, where the 0-shot model often provided a made-up citation (XYZ et al.), possibly from background knowledge. Only 6 cases of hallucination were found for GPT-4 plan compared to 16 instances for 0-shot vanilla GPT-4 model.

\section{Conclusions \& Answering: Are We There Yet?}
\label{section:conclusion}
This work discusses, establishes and evaluates a pipeline to help people write literature reviews. We first identified some challenges of evaluating such systems when LLMs are constantly updated based on training on new data, which may contain recent papers found online. To address these issues, we propose and implement a rolling evaluation procedure that focuses on recent arXiv papers, and we collect several evaluation datasets in this manner. 

Our experiments show that LLMs have significant potential for writing literature reviews, especially when the task is decomposed into these smaller and simpler sub-tasks that are within reach of LLMs, namely through the use of LLM-generated keyword search and embedding-based search for relevant prior work. Notably, our experiments indicate that both debate prompting and debate arguments that use attribution based on citing extracted content from source material improve LLM re-ranking results. The most powerful LLMs evaluated in our studies exhibit extremely promising paper re-ranking abilities as well as promising literature review generation results. Importantly, LLM hallucinations can be substantially reduced using our proposed plan-based prompting and retrieval augmented generation techniques. 

Our evaluation also reveals clear challenges: 1) retrieving all relevant papers consistent with a given human-generated literature review will require new querying strategies; 2) hallucinations can be significantly reduced using plan-based prompting, but our approach does not completely eliminate hallucinations.

So, \textit{are we there yet? Not quite—but we are getting closer.} The landscape of AI-assisted research exploded in 2025, with tools like DeepResearch~\citep{openai2025deepresearch}, AI Co-Scientist~\citep{gottweis2025towards}, and ScholarQA~\citep{scholarqa2025} demonstrating remarkable improvements in literature review generation, citation accuracy, and retrieval strategies.
Moreover, to accompany our scientific work here, and our older work on this theme \citep{agarwal2024litllm}, we have build a full working demo based on our proposed retrieval and generation pipeline (see Figure \ref{fig:demo} in the Appendix for a screenshot), which we will release to the community. We hope that authors can use this demonstration system to better understand how these techniques---and future alternatives---can be most helpful for assistance in literature review generation.



\textbf{Limitations and Future Work.} Because of the low coverage for retrieval, we evaluate different components independently. During generation, this strategy assumes that we already have filtered relevant papers corresponding to the main paper. In the future, we would like to improve the search for relevant work using embedding-based models to get better coverage and, thus, the ability to evaluate the system end-to-end. Our retrieval component currently suffers from surface-level information about the papers, while in practice, authors frame search keywords based on information (such as underlying datasets) that might not be present in the abstract. 
Another issue stems from the retrieval evaluation setup based on coverage related to the ground truth papers, where the authors might have different biases. 
%
We also acknowledge that including more of the paper, like introduction and methodology, might help improve the set of initial candidate papers; however, using additional sections could inadvertently allow the model to detect explicit citations or references, essentially ``cheating'' by using these as hints to retrieve specific papers. By focusing on the abstract alone, we maintain a more controlled setup that reflects a realistic, early-stage research scenario.
It is important to highlight that while we operate in this abstract-only setup, our pipeline is designed to be flexible and interactive. As the paper matures and more content becomes available, researchers can provide additional context to the pipeline to improve retrieval accuracy. This adaptability ensures that our approach remains relevant and effective throughout different stages of the research process, allowing for incremental refinement of related work as drafts evolve.






\appendix

\bibliography{main}

\begin{thebibliography}{85}
\providecommand{\natexlab}[1]{#1}
\providecommand{\url}[1]{\texttt{#1}}
\expandafter\ifx\csname urlstyle\endcsname\relax
  \providecommand{\doi}[1]{doi: #1}\else
  \providecommand{\doi}{doi: \begingroup \urlstyle{rm}\Url}\fi

\bibitem[Adebayo et~al.(2018)Adebayo, Gilmer, Muelly, Goodfellow, Hardt, and Kim]{adebayo2018sanity}
Julius Adebayo, Justin Gilmer, Michael Muelly, Ian Goodfellow, Moritz Hardt, and Been Kim.
\newblock Sanity checks for saliency maps.
\newblock In \emph{Advances in Neural Information Processing Systems}, pp.\  9505--9515, 2018.

\bibitem[Agarwal et~al.(2024)Agarwal, Laradji, Charlin, and Pal]{agarwal2024litllm}
Shubham Agarwal, Issam~H Laradji, Laurent Charlin, and Christopher Pal.
\newblock Litllm: A toolkit for scientific literature review.
\newblock \emph{arXiv preprint arXiv:2402.01788}, 2024.

\bibitem[AllenAI(2025)]{scholarqa2025}
AllenAI.
\newblock Introducing ai2 scholarqa, 2025.
\newblock URL \url{https://allenai.org/blog/ai2-scholarqa}.
\newblock Accessed: 2025-03-19.

\bibitem[Almazrouei et~al.(2023)Almazrouei, Alobeidli, Alshamsi, Cappelli, Cojocaru, Alhammadi, Daniele, Heslow, Launay, Malartic, Noune, Pannier, and Penedo]{almazrouei2023falcon}
Ebtesam Almazrouei, Hamza Alobeidli, Abdulaziz Alshamsi, Alessandro Cappelli, Ruxandra Cojocaru, Maitha Alhammadi, Mazzotta Daniele, Daniel Heslow, Julien Launay, Quentin Malartic, Badreddine Noune, Baptiste Pannier, and Guilherme Penedo.
\newblock The falcon series of language models: Towards open frontier models.
\newblock \emph{To appear}, 2023.

\bibitem[Athaluri et~al.(2023)Athaluri, Manthena, Kesapragada, Yarlagadda, Dave, and Duddumpudi]{Athaluri2023ExploringTB}
Sai~Anirudh Athaluri, Sandeep~Varma Manthena, V~S R Krishna~Manoj Kesapragada, Vineel Yarlagadda, Tirth Dave, and Rama Tulasi~Siri Duddumpudi.
\newblock Exploring the boundaries of reality: Investigating the phenomenon of artificial intelligence hallucination in scientific writing through chatgpt references.
\newblock \emph{Cureus}, 15, 2023.
\newblock URL \url{https://api.semanticscholar.org/CorpusID:258097853}.

\bibitem[Awadalla et~al.(2024)Awadalla, Xue, Lo, Shu, Lee, Guha, Jordan, Shen, Awadalla, Savarese, Xiong, Xu, Choi, and Schmidt]{awadalla2024mint1tscalingopensourcemultimodal}
Anas Awadalla, Le~Xue, Oscar Lo, Manli Shu, Hannah Lee, Etash~Kumar Guha, Matt Jordan, Sheng Shen, Mohamed Awadalla, Silvio Savarese, Caiming Xiong, Ran Xu, Yejin Choi, and Ludwig Schmidt.
\newblock Mint-1t: Scaling open-source multimodal data by 10x: A multimodal dataset with one trillion tokens, 2024.
\newblock URL \url{https://arxiv.org/abs/2406.11271}.

\bibitem[Beltagy et~al.(2019)Beltagy, Lo, and Cohan]{beltagy-etal-2019-scibert}
Iz~Beltagy, Kyle Lo, and Arman Cohan.
\newblock {S}ci{BERT}: A pretrained language model for scientific text.
\newblock In Kentaro Inui, Jing Jiang, Vincent Ng, and Xiaojun Wan (eds.), \emph{Proceedings of the 2019 Conference on Empirical Methods in Natural Language Processing and the 9th International Joint Conference on Natural Language Processing (EMNLP-IJCNLP)}, pp.\  3615--3620, Hong Kong, China, November 2019. Association for Computational Linguistics.
\newblock \doi{10.18653/v1/D19-1371}.
\newblock URL \url{https://aclanthology.org/D19-1371}.

\bibitem[Blecher et~al.(2023)Blecher, Cucurull, Scialom, and Stojnic]{blecher2023nougat}
Lukas Blecher, Guillem Cucurull, Thomas Scialom, and Robert Stojnic.
\newblock Nougat: Neural optical understanding for academic documents, 2023.

\bibitem[Brown et~al.(2020)Brown, Mann, Ryder, Subbiah, Kaplan, Dhariwal, Neelakantan, Shyam, Sastry, Askell, Agarwal, Herbert-Voss, Krueger, Henighan, Child, Ramesh, Ziegler, Wu, Winter, Hesse, Chen, Sigler, Litwin, Gray, Chess, Clark, Berner, McCandlish, Radford, Sutskever, and Amodei]{brown2020gpt3}
Tom~B. Brown, Benjamin Mann, Nick Ryder, Melanie Subbiah, Jared Kaplan, Prafulla Dhariwal, Arvind Neelakantan, Pranav Shyam, Girish Sastry, Amanda Askell, Sandhini Agarwal, Ariel Herbert-Voss, Gretchen Krueger, Tom Henighan, Rewon Child, Aditya Ramesh, Daniel~M. Ziegler, Jeffrey Wu, Clemens Winter, Christopher Hesse, Mark Chen, Eric Sigler, Mateusz Litwin, Scott Gray, Benjamin Chess, Jack Clark, Christopher Berner, Sam McCandlish, Alec Radford, Ilya Sutskever, and Dario Amodei.
\newblock Language models are few-shot learners, 2020.
\newblock URL \url{https://arxiv.org/abs/2005.14165}.

\bibitem[Cachola et~al.(2020)Cachola, Lo, Cohan, and Weld]{cachola-etal-2020-tldr}
Isabel Cachola, Kyle Lo, Arman Cohan, and Daniel Weld.
\newblock {TLDR}: Extreme summarization of scientific documents.
\newblock In \emph{Findings of the Association for Computational Linguistics: EMNLP 2020}, pp.\  4766--4777, Online, November 2020. Association for Computational Linguistics.
\newblock \doi{10.18653/v1/2020.findings-emnlp.428}.
\newblock URL \url{https://aclanthology.org/2020.findings-emnlp.428}.

\bibitem[Chen et~al.(2021{\natexlab{a}})Chen, Takamura, and Nakayama]{chen-etal-2021-scixgen-scientific}
Hong Chen, Hiroya Takamura, and Hideki Nakayama.
\newblock {S}ci{XG}en: A scientific paper dataset for context-aware text generation.
\newblock In \emph{Findings of the Association for Computational Linguistics: EMNLP 2021}, pp.\  1483--1492, Punta Cana, Dominican Republic, November 2021{\natexlab{a}}. Association for Computational Linguistics.
\newblock \doi{10.18653/v1/2021.findings-emnlp.128}.
\newblock URL \url{https://aclanthology.org/2021.findings-emnlp.128}.

\bibitem[Chen et~al.(2021{\natexlab{b}})Chen, Alamro, Li, Gao, Zhang, Zhao, and Yan]{chen-etal-2021-capturing}
Xiuying Chen, Hind Alamro, Mingzhe Li, Shen Gao, Xiangliang Zhang, Dongyan Zhao, and Rui Yan.
\newblock Capturing relations between scientific papers: An abstractive model for related work section generation.
\newblock In \emph{Proceedings of the 59th Annual Meeting of the Association for Computational Linguistics and the 11th International Joint Conference on Natural Language Processing (Volume 1: Long Papers)}, pp.\  6068--6077, Online, August 2021{\natexlab{b}}. Association for Computational Linguistics.
\newblock \doi{10.18653/v1/2021.acl-long.473}.
\newblock URL \url{https://aclanthology.org/2021.acl-long.473}.

\bibitem[Chung et~al.(2022)Chung, Hou, Longpre, Zoph, Tay, Fedus, Li, Wang, Dehghani, Brahma, et~al.]{chung2022scaling}
Hyung~Won Chung, Le~Hou, Shayne Longpre, Barret Zoph, Yi~Tay, William Fedus, Eric Li, Xuezhi Wang, Mostafa Dehghani, Siddhartha Brahma, et~al.
\newblock Scaling instruction-finetuned language models.
\newblock \emph{arXiv preprint arXiv:2210.11416}, 2022.

\bibitem[Cohan et~al.(2018)Cohan, Dernoncourt, Kim, Bui, Kim, Chang, and Goharian]{cohan-etal-2018-discourse}
Arman Cohan, Franck Dernoncourt, Doo~Soon Kim, Trung Bui, Seokhwan Kim, Walter Chang, and Nazli Goharian.
\newblock A discourse-aware attention model for abstractive summarization of long documents.
\newblock In \emph{Proceedings of the 2018 Conference of the North {A}merican Chapter of the Association for Computational Linguistics: Human Language Technologies, Volume 2 (Short Papers)}, pp.\  615--621, New Orleans, Louisiana, June 2018. Association for Computational Linguistics.
\newblock \doi{10.18653/v1/N18-2097}.
\newblock URL \url{https://aclanthology.org/N18-2097}.

\bibitem[Cohan et~al.(2020)Cohan, Feldman, Beltagy, Downey, and Weld]{cohan2020specter}
Arman Cohan, Sergey Feldman, Iz~Beltagy, Doug Downey, and Daniel~S Weld.
\newblock Specter: Document-level representation learning using citation-informed transformers.
\newblock \emph{arXiv preprint arXiv:2004.07180}, 2020.

\bibitem[Cohen-Wang et~al.(2024)Cohen-Wang, Shah, Georgiev, and Madry]{cohenwang2024contextcite}
Benjamin Cohen-Wang, Harshay Shah, Kristian Georgiev, and Aleksander Madry.
\newblock Contextcite: Attributing model generation to context, 2024.
\newblock URL \url{https://arxiv.org/abs/2409.00729}.

\bibitem[Devlin et~al.(2018)Devlin, Chang, Lee, and Toutanova]{devlin2018bert}
Jacob Devlin, Ming-Wei Chang, Kenton Lee, and Kristina Toutanova.
\newblock Bert: Pre-training of deep bidirectional transformers for language understanding.
\newblock \emph{arXiv preprint arXiv:1810.04805}, 2018.

\bibitem[Erkan \& Radev(2004)Erkan and Radev]{erkan2004lexrank}
G{\"u}nes Erkan and Dragomir~R Radev.
\newblock Lexrank: Graph-based lexical centrality as salience in text summarization.
\newblock \emph{Journal of artificial intelligence research}, 22:\penalty0 457--479, 2004.

\bibitem[Fong \& Vedaldi(2017)Fong and Vedaldi]{fong2017interpretable}
Ruth~C. Fong and Andrea Vedaldi.
\newblock Interpretable explanations of black boxes by meaningful perturbation.
\newblock In \emph{Proceedings of the IEEE International Conference on Computer Vision}, pp.\  3429--3437, 2017.

\bibitem[Funkquist et~al.(2022)Funkquist, Kuznetsov, Hou, and Gurevych]{funkquist2022citebench}
Martin Funkquist, Ilia Kuznetsov, Yufang Hou, and Iryna Gurevych.
\newblock Citebench: A benchmark for scientific citation text generation.
\newblock \emph{arXiv preprint arXiv:2212.09577}, 2022.

\bibitem[Gao et~al.(2023)Gao, Yen, Yu, and Chen]{gao2023enabling}
Tianyu Gao, Howard Yen, Jiatong Yu, and Danqi Chen.
\newblock Enabling large language models to generate text with citations.
\newblock \emph{arXiv preprint arXiv:2305.14627}, 2023.

\bibitem[Gehrmann et~al.(2018)Gehrmann, Dai, Elder, and Rush]{gehrmann-etal-2018-end}
Sebastian Gehrmann, Falcon Dai, Henry Elder, and Alexander Rush.
\newblock End-to-end content and plan selection for data-to-text generation.
\newblock In \emph{Proceedings of the 11th International Conference on Natural Language Generation}, pp.\  46--56, Tilburg University, The Netherlands, November 2018. Association for Computational Linguistics.
\newblock \doi{10.18653/v1/W18-6505}.
\newblock URL \url{https://aclanthology.org/W18-6505}.

\bibitem[Ghorbani et~al.(2019)Ghorbani, Abid, and Zou]{ghorbani2019interpretation}
Amirata Ghorbani, Abubakar Abid, and James Zou.
\newblock Interpretation of neural networks is fragile.
\newblock In \emph{Proceedings of the AAAI Conference on Artificial Intelligence}, volume~33, pp.\  3681--3688, 2019.

\bibitem[Gottweis et~al.(2025)Gottweis, Weng, Daryin, Tu, Palepu, Sirkovic, Myaskovsky, Weissenberger, Rong, Tanno, et~al.]{gottweis2025towards}
Juraj Gottweis, Wei-Hung Weng, Alexander Daryin, Tao Tu, Anil Palepu, Petar Sirkovic, Artiom Myaskovsky, Felix Weissenberger, Keran Rong, Ryutaro Tanno, et~al.
\newblock Towards an ai co-scientist.
\newblock \emph{arXiv preprint arXiv:2502.18864}, 2025.

\bibitem[Guo et~al.(2022)Guo, Ainslie, Uthus, Ontanon, Ni, Sung, and Yang]{guo-etal-2022-longt5}
Mandy Guo, Joshua Ainslie, David Uthus, Santiago Ontanon, Jianmo Ni, Yun-Hsuan Sung, and Yinfei Yang.
\newblock {L}ong{T}5: {E}fficient text-to-text transformer for long sequences.
\newblock In \emph{Findings of the Association for Computational Linguistics: NAACL 2022}, pp.\  724--736, Seattle, United States, July 2022. Association for Computational Linguistics.
\newblock \doi{10.18653/v1/2022.findings-naacl.55}.
\newblock URL \url{https://aclanthology.org/2022.findings-naacl.55}.

\bibitem[Kadous(2023)]{kadous_2023}
M~Waleed Kadous.
\newblock Llama 2 is about as factually accurate as gpt-4 for summaries and is 30x cheaper, Aug 2023.
\newblock URL \url{https://www.anyscale.com/blog/llama-2-is-about-as-factually-accurate-as-gpt-4-for-summaries-and-is-30x-cheaper}.

\bibitem[Keeling \& Street(2024)Keeling and Street]{keeling2024attribution}
Geoff Keeling and Winnie Street.
\newblock On the attribution of confidence to large language models, 2024.
\newblock URL \url{https://arxiv.org/abs/2407.08388}.

\bibitem[Kindermans et~al.(2019)Kindermans, Hooker, Adebayo, Alber, Sch{\"u}tt, D{\"a}hne, Erhan, and Kim]{kindermans2019reliability}
Pieter-Jan Kindermans, Sara Hooker, Julius Adebayo, Maximilian Alber, Kristof~T. Sch{\"u}tt, Sven D{\"a}hne, Dumitru Erhan, and Been Kim.
\newblock The (un) reliability of saliency methods.
\newblock \emph{Explainable AI: Interpreting, Explaining and Visualizing Deep Learning}, pp.\  267--280, 2019.

\bibitem[Kinney et~al.(2023)Kinney, Anastasiades, Authur, Beltagy, Bragg, Buraczynski, Cachola, Candra, Chandrasekhar, Cohan, et~al.]{kinney2023semantic}
Rodney Kinney, Chloe Anastasiades, Russell Authur, Iz~Beltagy, Jonathan Bragg, Alexandra Buraczynski, Isabel Cachola, Stefan Candra, Yoganand Chandrasekhar, Arman Cohan, et~al.
\newblock The semantic scholar open data platform.
\newblock \emph{arXiv preprint arXiv:2301.10140}, 2023.

\bibitem[Kojima et~al.(2022)Kojima, Gu, Reid, Matsuo, and Iwasawa]{kojima2022large}
Takeshi Kojima, Shixiang~Shane Gu, Machel Reid, Yutaka Matsuo, and Yusuke Iwasawa.
\newblock Large language models are zero-shot reasoners.
\newblock \emph{Advances in neural information processing systems}, 35:\penalty0 22199--22213, 2022.

\bibitem[Lachenbruch(2014)]{lachenbruch2014mcnemar}
Peter~A Lachenbruch.
\newblock Mcnemar test.
\newblock \emph{Wiley StatsRef: Statistics Reference Online}, 2014.

\bibitem[Lewis et~al.(2020)Lewis, Perez, Piktus, Petroni, Karpukhin, Goyal, K{\"u}ttler, Lewis, Yih, Rockt{\"a}schel, et~al.]{lewis2020retrieval}
Patrick Lewis, Ethan Perez, Aleksandra Piktus, Fabio Petroni, Vladimir Karpukhin, Naman Goyal, Heinrich K{\"u}ttler, Mike Lewis, Wen-tau Yih, Tim Rockt{\"a}schel, et~al.
\newblock Retrieval-augmented generation for knowledge-intensive nlp tasks.
\newblock \emph{Advances in Neural Information Processing Systems}, 33:\penalty0 9459--9474, 2020.

\bibitem[Li et~al.(2023)Li, Sun, Hu, Liu, Chen, Hu, Wu, and Zhang]{li2023surveylargelanguagemodels}
Dongfang Li, Zetian Sun, Xinshuo Hu, Zhenyu Liu, Ziyang Chen, Baotian Hu, Aiguo Wu, and Min Zhang.
\newblock A survey of large language models attribution, 2023.
\newblock URL \url{https://arxiv.org/abs/2311.03731}.

\bibitem[Li et~al.(2016)Li, Chen, Hovy, and Jurafsky]{li2016understanding}
Jiwei Li, Xinlei Chen, Eduard Hovy, and Dan Jurafsky.
\newblock Visualizing and understanding neural models in nlp.
\newblock In \emph{Proceedings of the 2016 Conference of the North American Chapter of the Association for Computational Linguistics}, pp.\  681--691. Association for Computational Linguistics, 2016.

\bibitem[Li et~al.(2022)Li, Mandal, and Ouyang]{li-etal-2022-corwa}
Xiangci Li, Biswadip Mandal, and Jessica Ouyang.
\newblock {CORWA}: A citation-oriented related work annotation dataset.
\newblock In \emph{Proceedings of the 2022 Conference of the North American Chapter of the Association for Computational Linguistics: Human Language Technologies}, pp.\  5426--5440, Seattle, United States, July 2022. Association for Computational Linguistics.
\newblock \doi{10.18653/v1/2022.naacl-main.397}.
\newblock URL \url{https://aclanthology.org/2022.naacl-main.397}.

\bibitem[Lin(2004)]{lin-2004-rouge}
Chin-Yew Lin.
\newblock {ROUGE}: A package for automatic evaluation of summaries.
\newblock In \emph{Text Summarization Branches Out}, pp.\  74--81, Barcelona, Spain, July 2004. Association for Computational Linguistics.
\newblock URL \url{https://aclanthology.org/W04-1013}.

\bibitem[Liu et~al.(2023{\natexlab{a}})Liu, Cao, Yang, and Wen]{liu2023generating}
Shuaiqi Liu, Jiannong Cao, Ruosong Yang, and Zhiyuan Wen.
\newblock Generating a structured summary of numerous academic papers: Dataset and method.
\newblock \emph{arXiv preprint arXiv:2302.04580}, 2023{\natexlab{a}}.

\bibitem[Liu \& Lapata(2019)Liu and Lapata]{liu2019hiersumm}
Yang Liu and Mirella Lapata.
\newblock Hierarchical transformers for multi-document summarization.
\newblock In \emph{Proceedings of the 57th Annual Meeting of the Association for Computational Linguistics}, pp.\  5070--5081, 2019.

\bibitem[Liu et~al.(2023{\natexlab{b}})Liu, Iter, Xu, Wang, Xu, and Zhu]{liu-etal-2023-g}
Yang Liu, Dan Iter, Yichong Xu, Shuohang Wang, Ruochen Xu, and Chenguang Zhu.
\newblock {G}-eval: {NLG} evaluation using gpt-4 with better human alignment.
\newblock In Houda Bouamor, Juan Pino, and Kalika Bali (eds.), \emph{Proceedings of the 2023 Conference on Empirical Methods in Natural Language Processing}, pp.\  2511--2522, Singapore, December 2023{\natexlab{b}}. Association for Computational Linguistics.
\newblock \doi{10.18653/v1/2023.emnlp-main.153}.
\newblock URL \url{https://aclanthology.org/2023.emnlp-main.153/}.

\bibitem[Lo et~al.(2020)Lo, Wang, Neumann, Kinney, and Weld]{lo-etal-2020-s2orc}
Kyle Lo, Lucy~Lu Wang, Mark Neumann, Rodney Kinney, and Daniel Weld.
\newblock {S}2{ORC}: The semantic scholar open research corpus.
\newblock In \emph{Proceedings of the 58th Annual Meeting of the Association for Computational Linguistics}, pp.\  4969--4983, Online, July 2020. Association for Computational Linguistics.
\newblock \doi{10.18653/v1/2020.acl-main.447}.
\newblock URL \url{https://aclanthology.org/2020.acl-main.447}.

\bibitem[Lopez(2023)]{lopez_grobid_2023}
Patrice Lopez.
\newblock {GROBID}, February 2023.
\newblock URL \url{https://github.com/kermitt2/grobid}.
\newblock original-date: 2012-09-13T15:48:54Z.

\bibitem[Lu et~al.(2020)Lu, Dong, and Charlin]{lu-etal-2020-multi-xscience}
Yao Lu, Yue Dong, and Laurent Charlin.
\newblock Multi-{XS}cience: A large-scale dataset for extreme multi-document summarization of scientific articles.
\newblock In \emph{Proceedings of the 2020 Conference on Empirical Methods in Natural Language Processing (EMNLP)}, pp.\  8068--8074. Association for Computational Linguistics, November 2020.
\newblock \doi{10.18653/v1/2020.emnlp-main.648}.
\newblock URL \url{https://aclanthology.org/2020.emnlp-main.648}.

\bibitem[Ma et~al.(2023)Ma, Zhang, Pradeep, and Lin]{ma2023zero}
Xueguang Ma, Xinyu Zhang, Ronak Pradeep, and Jimmy Lin.
\newblock Zero-shot listwise document reranking with a large language model.
\newblock \emph{arXiv preprint arXiv:2305.02156}, 2023.

\bibitem[Mihalcea \& Tarau(2004)Mihalcea and Tarau]{mihalcea2004textrank}
Rada Mihalcea and Paul Tarau.
\newblock Textrank: Bringing order into text.
\newblock In \emph{Proceedings of the 2004 conference on empirical methods in natural language processing}, pp.\  404--411, 2004.

\bibitem[Nguyen et~al.(2023)Nguyen, Scialom, Piwowarski, and Staiano]{nguyen-etal-2023-loralay}
Laura Nguyen, Thomas Scialom, Benjamin Piwowarski, and Jacopo Staiano.
\newblock {L}o{R}a{L}ay: A multilingual and multimodal dataset for long range and layout-aware summarization.
\newblock In \emph{Proceedings of the 17th Conference of the European Chapter of the Association for Computational Linguistics}, pp.\  636--651, Dubrovnik, Croatia, May 2023. Association for Computational Linguistics.
\newblock \doi{10.18653/v1/2023.eacl-main.46}.
\newblock URL \url{https://aclanthology.org/2023.eacl-main.46}.

\bibitem[OpenAI(2023)]{openai2023gpt}
OpenAI.
\newblock {GPT-4} technical report.
\newblock \emph{arXiv}, 2023.

\bibitem[OpenAI(2025)]{openai2025deepresearch}
OpenAI.
\newblock Introducing deep research, 2025.
\newblock URL \url{https://openai.com/index/introducing-deep-research/}.
\newblock Accessed: 2025-03-19.

\bibitem[Paszke et~al.(2017)Paszke, Gross, Chintala, Chanan, Yang, DeVito, Lin, Desmaison, Antiga, and Lerer]{paszke2017automatic}
Adam Paszke, Sam Gross, Soumith Chintala, Gregory Chanan, Edward Yang, Zachary DeVito, Zeming Lin, Alban Desmaison, Luca Antiga, and Adam Lerer.
\newblock Automatic differentiation in {PyTorch}.
\newblock In \emph{NeurIPS-W}, 2017.
\newblock URL \url{https://openreview.net/forum?id=BJJsrmfCZ}.

\bibitem[Pilault et~al.(2020)Pilault, Li, Subramanian, and Pal]{pilault-etal-2020-extractive}
Jonathan Pilault, Raymond Li, Sandeep Subramanian, and Chris Pal.
\newblock On extractive and abstractive neural document summarization with transformer language models.
\newblock In \emph{Proceedings of the 2020 Conference on Empirical Methods in Natural Language Processing (EMNLP)}, pp.\  9308--9319, Online, November 2020. Association for Computational Linguistics.
\newblock \doi{10.18653/v1/2020.emnlp-main.748}.
\newblock URL \url{https://aclanthology.org/2020.emnlp-main.748}.

\bibitem[Pradeep et~al.(2023{\natexlab{a}})Pradeep, Sharifymoghaddam, and Lin]{pradeep2023rankvicuna}
Ronak Pradeep, Sahel Sharifymoghaddam, and Jimmy Lin.
\newblock Rankvicuna: Zero-shot listwise document reranking with open-source large language models.
\newblock \emph{arXiv preprint arXiv:2309.15088}, 2023{\natexlab{a}}.

\bibitem[Pradeep et~al.(2023{\natexlab{b}})Pradeep, Sharifymoghaddam, and Lin]{pradeep2023rankzephyr}
Ronak Pradeep, Sahel Sharifymoghaddam, and Jimmy Lin.
\newblock Rankzephyr: Effective and robust zero-shot listwise reranking is a breeze!
\newblock \emph{arXiv preprint arXiv:2312.02724}, 2023{\natexlab{b}}.

\bibitem[Priem et~al.(2022)Priem, Piwowar, and Orr]{priem2022openalex}
Jason Priem, Heather Piwowar, and Richard Orr.
\newblock Openalex: A fully-open index of scholarly works, authors, venues, institutions, and concepts.
\newblock \emph{arXiv preprint arXiv:2205.01833}, 2022.

\bibitem[Puduppully \& Lapata(2021)Puduppully and Lapata]{puduppully-2021-macro}
Ratish Puduppully and Mirella Lapata.
\newblock Data-to-text generation with macro planning.
\newblock \emph{Transactions of the Association for Computational Linguistics}, 9:\penalty0 510--527, 2021.
\newblock URL \url{https://direct.mit.edu/tacl/article/doi/10.1162/tacl_a_00381/101876/Data-to-text-Generation-with-Macro-Planning}.

\bibitem[Puduppully et~al.(2019)Puduppully, Dong, and Lapata]{puduppully-2019-data}
Ratish Puduppully, Li~Dong, and Mirella Lapata.
\newblock Data-to-text generation with content selection and planning.
\newblock In \emph{The Thirty-Third {AAAI} Conference on Artificial Intelligence, {AAAI} 2019, The Thirty-First Innovative Applications of Artificial Intelligence Conference, {IAAI} 2019, The Ninth {AAAI} Symposium on Educational Advances in Artificial Intelligence, {EAAI} 2019, Honolulu, Hawaii, USA, January 27 - February 1, 2019}, pp.\  6908--6915. {AAAI} Press, 2019.
\newblock \doi{10.1609/aaai.v33i01.33016908}.
\newblock URL \url{https://doi.org/10.1609/aaai.v33i01.33016908}.

\bibitem[Raffel et~al.(2020)Raffel, Shazeer, Roberts, Lee, Narang, Matena, Zhou, Li, and Liu]{raffel2020exploring}
Colin Raffel, Noam Shazeer, Adam Roberts, Katherine Lee, Sharan Narang, Michael Matena, Yanqi Zhou, Wei Li, and Peter~J Liu.
\newblock Exploring the limits of transfer learning with a unified text-to-text transformer.
\newblock \emph{The Journal of Machine Learning Research}, 21\penalty0 (1):\penalty0 5485--5551, 2020.

\bibitem[Rahaman et~al.(2024)Rahaman, Weiss, W{\"u}thrich, Bengio, Li, Pal, and Sch{\"o}lkopf]{rahaman2024language}
Nasim Rahaman, Martin Weiss, Manuel W{\"u}thrich, Yoshua Bengio, Li~Erran Li, Chris Pal, and Bernhard Sch{\"o}lkopf.
\newblock Language models can reduce asymmetry in information markets.
\newblock \emph{arXiv preprint arXiv:2403.14443}, 2024.

\bibitem[Reimers \& Gurevych(2019)Reimers and Gurevych]{reimers-sbert}
Nils Reimers and Iryna Gurevych.
\newblock Sentence-{BERT}: Sentence embeddings using {S}iamese {BERT}-networks.
\newblock In Kentaro Inui, Jing Jiang, Vincent Ng, and Xiaojun Wan (eds.), \emph{Proceedings of the 2019 Conference on Empirical Methods in Natural Language Processing and the 9th International Joint Conference on Natural Language Processing (EMNLP-IJCNLP)}, pp.\  3982--3992, Hong Kong, China, November 2019. Association for Computational Linguistics.
\newblock \doi{10.18653/v1/D19-1410}.
\newblock URL \url{https://aclanthology.org/D19-1410/}.

\bibitem[Reiter \& Dale(1997)Reiter and Dale]{reiter1997building}
Ehud Reiter and Robert Dale.
\newblock Building applied natural language generation systems.
\newblock \emph{Natural Language Engineering}, 3\penalty0 (1):\penalty0 57--87, 1997.

\bibitem[Ribeiro et~al.(2016)Ribeiro, Singh, and Guestrin]{ribeiro2016should}
Marco~Tulio Ribeiro, Sameer Singh, and Carlos Guestrin.
\newblock \textquotedblleft why should i trust you?\textquotedblright: Explaining the predictions of any classifier.
\newblock In \emph{Proceedings of the 22nd ACM SIGKDD International Conference on Knowledge Discovery and Data Mining}, pp.\  1135--1144. ACM, 2016.

\bibitem[Rodriguez et~al.(2024{\natexlab{a}})Rodriguez, Jian, Panigrahi, Zhang, Feizi, Puri, Kalkunte, Savard, Masry, Nayak, Awal, Massoud, Abaskohi, Li, Wang, Noël, Richter, Vadacchino, Agarwal, Biswas, Shanian, Zhang, Bolger, MacDonald, Fauvel, Tejaswi, Sunkara, Monteiro, Dvijotham, Scholak, Chapados, Kharagani, Hughes, Özsu, Reddy, Pedersoli, Bengio, Pal, Laradji, Gella, Taslakian, Vazquez, and Rajeswar]{rodriguez2024bigdocs}
Juan Rodriguez, Xiangru Jian, Siba~Smarak Panigrahi, Tianyu Zhang, Aarash Feizi, Abhay Puri, Akshay Kalkunte, François Savard, Ahmed Masry, Shravan Nayak, Rabiul Awal, Mahsa Massoud, Amirhossein Abaskohi, Zichao Li, Suyuchen Wang, Pierre-André Noël, Mats~Leon Richter, Saverio Vadacchino, Shubbam Agarwal, Sanket Biswas, Sara Shanian, Ying Zhang, Noah Bolger, Kurt MacDonald, Simon Fauvel, Sathwik Tejaswi, Srinivas Sunkara, Joao Monteiro, Krishnamurthy~DJ Dvijotham, Torsten Scholak, Nicolas Chapados, Sepideh Kharagani, Sean Hughes, M.~Özsu, Siva Reddy, Marco Pedersoli, Yoshua Bengio, Christopher Pal, Issam Laradji, Spandanna Gella, Perouz Taslakian, David Vazquez, and Sai Rajeswar.
\newblock Bigdocs: An open and permissively-licensed dataset for training multimodal models on document and code tasks, 2024{\natexlab{a}}.
\newblock URL \url{https://arxiv.org/abs/2412.04626}.

\bibitem[Rodriguez et~al.(2024{\natexlab{b}})Rodriguez, Puri, Agarwal, Laradji, Rodriguez, Rajeswar, Vazquez, Pal, and Pedersoli]{rodriguez2024starvectorgeneratingscalablevector}
Juan~A. Rodriguez, Abhay Puri, Shubham Agarwal, Issam~H. Laradji, Pau Rodriguez, Sai Rajeswar, David Vazquez, Christopher Pal, and Marco Pedersoli.
\newblock Starvector: Generating scalable vector graphics code from images and text, 2024{\natexlab{b}}.
\newblock URL \url{https://arxiv.org/abs/2312.11556}.

\bibitem[Saier \& F{\"{a}}rber(2020)Saier and F{\"{a}}rber]{Saier2020unarXive}
Tarek Saier and Michael F{\"{a}}rber.
\newblock {unarXive: A Large Scholarly Data Set with Publications’ Full-Text, Annotated In-Text Citations, and Links to Metadata}.
\newblock \emph{Scientometrics}, 125\penalty0 (3):\penalty0 3085--3108, December 2020.
\newblock ISSN 1588-2861.
\newblock \doi{10.1007/s11192-020-03382-z}.

\bibitem[Saier et~al.(2023)Saier, Krause, and F\"{a}rber]{saier2023unarxive}
Tarek Saier, Johan Krause, and Michael F\"{a}rber.
\newblock {unarXive 2022: All arXiv Publications Pre-Processed for NLP, Including Structured Full-Text and Citation Network}.
\newblock In \emph{Proceedings of the 23rd ACM/IEEE Joint Conference on Digital Libraries}, JCDL '23, 2023.

\bibitem[See et~al.(2017)See, Liu, and Manning]{see2017pointer}
Abigail See, Peter~J Liu, and Christopher~D Manning.
\newblock Get to the point: Summarization with pointer-generator networks.
\newblock In \emph{Proceedings of the 55th Annual Meeting of the Association for Computational Linguistics}, pp.\  1073--1083, 2017.

\bibitem[Shrikumar et~al.(2017)Shrikumar, Greenside, and Kundaje]{shrikumar2017learning}
Avanti Shrikumar, Peyton Greenside, and Anshul Kundaje.
\newblock Learning important features through propagating activation differences.
\newblock In \emph{International Conference on Machine Learning}, pp.\  3145--3153. PMLR, 2017.

\bibitem[Simonyan et~al.(2014)Simonyan, Vedaldi, and Zisserman]{simonyan2014deep}
Karen Simonyan, Andrea Vedaldi, and Andrew Zisserman.
\newblock Deep inside convolutional networks: Visualising image classification models and saliency maps.
\newblock 2014.
\newblock URL \url{https://arxiv.org/abs/1312.6034}.

\bibitem[Singh et~al.(2022)Singh, D'Arcy, Cohan, Downey, and Feldman]{singh2022scirepeval}
Amanpreet Singh, Mike D'Arcy, Arman Cohan, Doug Downey, and Sergey Feldman.
\newblock {SciRepEval}: A multi-format benchmark for scientific document representations.
\newblock \emph{arXiv preprint arXiv:2211.13308}, 2022.

\bibitem[Sinha et~al.(2015)Sinha, Shen, Song, Ma, Eide, Hsu, and Wang]{sinha2015overview}
Arnab Sinha, Zhihong Shen, Yang Song, Hao Ma, Darrin Eide, Bo-June Hsu, and Kuansan Wang.
\newblock An overview of microsoft academic service (mas) and applications.
\newblock In \emph{Proceedings of the 24th international conference on world wide web}, pp.\  243--246, 2015.

\bibitem[Smilkov et~al.(2017)Smilkov, Thorat, Kim, Viégas, and Wattenberg]{smilkov2017smoothgrad}
Daniel Smilkov, Nikhil Thorat, Been Kim, Fernanda Viégas, and Martin Wattenberg.
\newblock Smoothgrad: Removing noise by adding noise.
\newblock 2017.
\newblock URL \url{https://arxiv.org/abs/1706.03825}.

\bibitem[Song et~al.(2023)Song, Wu, Washington, Sadler, Chao, and Su]{song2023llm}
Chan~Hee Song, Jiaman Wu, Clayton Washington, Brian~M Sadler, Wei-Lun Chao, and Yu~Su.
\newblock Llm-planner: Few-shot grounded planning for embodied agents with large language models.
\newblock In \emph{Proceedings of the IEEE/CVF International Conference on Computer Vision}, pp.\  2998--3009, 2023.

\bibitem[Stent et~al.(2004)Stent, Prasad, and Walker]{stent-etal-2004-trainable}
Amanda Stent, Rashmi Prasad, and Marilyn Walker.
\newblock Trainable sentence planning for complex information presentations in spoken dialog systems.
\newblock In \emph{Proceedings of the 42nd Annual Meeting of the Association for Computational Linguistics ({ACL}-04)}, pp.\  79--86, Barcelona, Spain, July 2004.
\newblock \doi{10.3115/1218955.1218966}.
\newblock URL \url{https://aclanthology.org/P04-1011}.

\bibitem[Su et~al.(2021)Su, Lu, Pan, Murtadha, Wen, and Liu]{su2021roformer}
Jianlin Su, Yu~Lu, Shengfeng Pan, Ahmed Murtadha, Bo~Wen, and Yunfeng Liu.
\newblock Roformer: Enhanced transformer with rotary position embedding.
\newblock \emph{arXiv preprint arXiv:2104.09864}, 2021.

\bibitem[Sun et~al.(2023)Sun, Yan, Ma, Ren, Yin, and Ren]{sun2023chatgpt}
Weiwei Sun, Lingyong Yan, Xinyu Ma, Pengjie Ren, Dawei Yin, and Zhaochun Ren.
\newblock Is {ChatGPT } good at search? investigating large language models as re-ranking agent.
\newblock \emph{arXiv preprint arXiv:2304.09542}, 2023.

\bibitem[Sundararajan et~al.(2017)Sundararajan, Taly, and Yan]{sundararajan2017axiomatic}
Mukund Sundararajan, Ankur Taly, and Qiqi Yan.
\newblock Axiomatic attribution for deep networks.
\newblock In \emph{International Conference on Machine Learning}, pp.\  3319--3328. PMLR, 2017.

\bibitem[Taylor et~al.(2022)Taylor, Kardas, Cucurull, Scialom, Hartshorn, Saravia, Poulton, Kerkez, and Stojnic]{taylor2022galactica}
Ross Taylor, Marcin Kardas, Guillem Cucurull, Thomas Scialom, Anthony Hartshorn, Elvis Saravia, Andrew Poulton, Viktor Kerkez, and Robert Stojnic.
\newblock Galactica: A large language model for science.
\newblock \emph{arXiv preprint arXiv:2211.09085}, 2022.

\bibitem[Touvron et~al.(2023)Touvron, Martin, Stone, Albert, Almahairi, Babaei, Bashlykov, Batra, Bhargava, Bhosale, et~al.]{touvron2023llama2}
Hugo Touvron, Louis Martin, Kevin Stone, Peter Albert, Amjad Almahairi, Yasmine Babaei, Nikolay Bashlykov, Soumya Batra, Prajjwal Bhargava, Shruti Bhosale, et~al.
\newblock Llama 2: Open foundation and fine-tuned chat models.
\newblock \emph{arXiv preprint arXiv:2307.09288}, 2023.

\bibitem[Walker et~al.(2007)Walker, Stent, Mairesse, and Prasad]{walker2007individual}
Marilyn~A Walker, Amanda Stent, Fran{\c{c}}ois Mairesse, and Rashmi Prasad.
\newblock Individual and domain adaptation in sentence planning for dialogue.
\newblock \emph{Journal of Artificial Intelligence Research}, 30:\penalty0 413--456, 2007.

\bibitem[Wang et~al.(2023)Wang, Xu, Lan, Hu, Lan, Lee, and Lim]{wang2023plan}
Lei Wang, Wanyu Xu, Yihuai Lan, Zhiqiang Hu, Yunshi Lan, Roy Ka-Wei Lee, and Ee-Peng Lim.
\newblock Plan-and-solve prompting: Improving zero-shot chain-of-thought reasoning by large language models.
\newblock \emph{arXiv preprint arXiv:2305.04091}, 2023.

\bibitem[Wolf et~al.(2019)Wolf, Debut, Sanh, Chaumond, Delangue, Moi, Cistac, Rault, Louf, Funtowicz, et~al.]{wolf2019huggingface}
Thomas Wolf, Lysandre Debut, Victor Sanh, Julien Chaumond, Clement Delangue, Anthony Moi, Pierric Cistac, Tim Rault, R{\'e}mi Louf, Morgan Funtowicz, et~al.
\newblock Huggingface's transformers: State-of-the-art natural language processing.
\newblock \emph{arXiv preprint arXiv:1910.03771}, 2019.

\bibitem[Xiao et~al.(2021)Xiao, Beltagy, Carenini, and Cohan]{xiao2021primera}
Wen Xiao, Iz~Beltagy, Giuseppe Carenini, and Arman Cohan.
\newblock Primera: Pyramid-based masked sentence pre-training for multi-document summarization.
\newblock \emph{arXiv preprint arXiv:2110.08499}, 2021.

\bibitem[Yang et~al.(2022)Yang, Tian, Peng, and Klein]{yang2022re3}
Kevin Yang, Yuandong Tian, Nanyun Peng, and Dan Klein.
\newblock Re3: Generating longer stories with recursive reprompting and revision.
\newblock \emph{arXiv preprint arXiv:2210.06774}, 2022.

\bibitem[Yue et~al.(2023)Yue, Wang, Chen, Zhang, Su, and Sun]{yue2023automatic}
Xiang Yue, Boshi Wang, Ziru Chen, Kai Zhang, Yu~Su, and Huan Sun.
\newblock Automatic evaluation of attribution by large language models, 2023.
\newblock URL \url{https://arxiv.org/abs/2305.06311}.

\bibitem[Zhang et~al.(2023)Zhang, Hofst{\"a}tter, Lewis, Tang, and Lin]{zhang2023rank}
Xinyu Zhang, Sebastian Hofst{\"a}tter, Patrick Lewis, Raphael Tang, and Jimmy Lin.
\newblock Rank-without-gpt: Building gpt-independent listwise rerankers on open-source large language models.
\newblock \emph{arXiv preprint arXiv:2312.02969}, 2023.

\bibitem[Zhou et~al.(2022)Zhou, Muresanu, Han, Paster, Pitis, Chan, and Ba]{zhou2022large}
Yongchao Zhou, Andrei~Ioan Muresanu, Ziwen Han, Keiran Paster, Silviu Pitis, Harris Chan, and Jimmy Ba.
\newblock Large language models are human-level prompt engineers.
\newblock \emph{arXiv preprint arXiv:2211.01910}, 2022.

\bibitem[Zhu et~al.(2023)Zhu, Feng, Feng, Wu, and Qin]{zhu2023hierarchical}
Kun Zhu, Xiaocheng Feng, Xiachong Feng, Yingsheng Wu, and Bing Qin.
\newblock Hierarchical catalogue generation for literature review: A benchmark.
\newblock \emph{arXiv preprint arXiv:2304.03512}, 2023.

\end{thebibliography}
\bibliographystyle{tmlr}

\newpage
\section*{Appendix}

\section{Ethics Statement}
The rapid advancements in LLMs and NLP technologies for scientific writing have led to the emergence of increasingly powerful systems such as DeepResearch, AI Co-Scientist, and ScholarQA.
These tools extend beyond earlier systems like Explainpaper and Writefull\footnote{\url{https://www.explainpaper.com/}, \url{https://x.writefull.com/}}, which assist in paper comprehension and abstract generation, and Scite\footnote{\url{https://scite.ai/}}, which helps with citation discovery.
As AI-powered tools become more deeply integrated into the scientific workflow, ethical considerations around their use continue to evolve. Many conferences, such as ICLR, have begun collecting statistics on authors' usage of LLMs for literature review generation and paraphrasing, and have issued guidelines on responsible usage.\footnote{ICLR'24 Large Language Models guidelines \url{https://iclr.cc/Conferences/2024/CallForPapers}}
While writing assistant technology could have great promise as an aide to scientists, we think their use should be disclosed to the reader. As such assistants become more powerful, they might be abused in certain contexts, for example, where students are supposed to create a literature review as a part of their learning process. The use of such tools might also be problematic as authors of scientific work should read the articles that they cite, and heavy reliance on such tools could lead to short-term gains at the cost of a deeper understanding of a subject over the longer term. Any commercially deployed systems authors use should also contain appropriate mechanisms to detect if words have been copied exactly from the source material and provide that content in a quoted style.
Additionally, as newer tools like DeepResearch, AI Co-Scientist, and ScholarQA continue to improve, it is crucial to assess their long-term impact on scientific research. The use of these tools should complement, rather than replace, human expertise in literature analysis. Finally, the rolling evaluations we present here do not involve training LLMs on arXiv papers. This mitigates concerns regarding the copyright status of arXiv papers and their use for LLM training. 




\section{New Datasets}
\label{appendix:datasets}
While there are datasets available for different tasks in academic literature (see Table \ref{table:tasks-datasets}), we use the Multi-XScience dataset ~\citep{lu-etal-2020-multi-xscience} for our experiments. Recent work ~\citep{chen-etal-2021-capturing,funkquist2022citebench} also focuses on related work generation and provides a similar dataset. 
As part of this work, we release two corpora:
1. We extend the Multi-XScience corpus to include the full text of research papers, and 2. We create a new test corpus, RollingEval-Aug, consisting of recent (August 2023) arXiv papers (with full content). 

\begin{table}[ht]
\centering
\begin{tabular}{l|l}
\toprule
\textbf{Dataset} & \textbf{Task} \\
\midrule
BigSurvey-MDS \citep{liu2023generating} & Survey Introduction \\
HiCaD \citep{zhu2023hierarchical} & Survey Catalogue \\
SciXGen \citep{chen-etal-2021-scixgen-scientific} & Context-aware text generation \\
CORWA \citep{li-etal-2022-corwa} & Citation Span Generation \\
TLDR \citep{cachola-etal-2020-tldr} & TLDR generation\\
Multi-XScience \cite{lu-etal-2020-multi-xscience} & Related Work Generation \\
\bottomrule
\end{tabular}
\caption{Different tasks for academic literature}
\label{table:tasks-datasets}
\end{table}

\textbf{Multi-XScience full text} We create these datasets based on the latest release (2023-09-12) of the S2ORC corpus\footnote{Dataset available at \url{http://api.semanticscholar.org/datasets/v1/}} ~\citep{lo-etal-2020-s2orc} available at the Semantic Scholar Open Data Platform ~\citep{kinney2023semantic}. The S2 Platform provides access to multiple datasets, including paper metadata, authors, S2AG (Semantic Scholar Academic Graph), paper embeddings, etc. While the `Papers' dataset consists of 200M+ metadata records, S2ORC consists of 11+M full-text publicly available records with annotations chunked into 30 files ($\sim$215G compressed json) where research documents are linked with arXiv and Microsoft Academic Graph (MAG) ~\citep{sinha2015overview} IDs, when available. This corpus provides full text of the research papers (parsed using a complex pipeline consisting of multiple LaTeX and PDF parsers such as GROBID ~\citep{lopez_grobid_2023} and in-house parsers.\footnote{\url{https://github.com/allenai/papermage}}). The full text is also aligned with annotation spans (character level on the full text), which identify sections, paragraphs, and other useful information. It also includes spans for citation mentions and the matching semantic corpus-based ID for bibliographical entries, making it easier to align with references compared to other academic datasets such as LoRaLay ~\citep{nguyen-etal-2023-loralay}, UnarXive ~\citep{Saier2020unarXive,saier2023unarxive}, etc. or relying on citation graphs like OpenAlex ~\citep{priem2022openalex}, next-generation PDF parsers ~\citep{blecher2023nougat} or other HTML webpages.\footnote{\url{https://ar5iv.labs.arxiv.org/} and \url{https://www.arxiv-vanity.com/}} For the Multi-XScience, we obtain the full text of papers for 85\% of records from the S2ORC data using the span annotations from the corpus aligned with citation information. 

\textbf{RollingEval datasets} Llama 2 was publicly released on 18th July 2023 and GPT-4 on 14 March 2023. Both provide limited information about their training corpus, and academic texts in the Multi-XScience may or may not have been part of their training data. To avoid overlap with the training data of these LLMs, we process a new dataset using papers posted after their release date. To do so, we first filter the papers published in August 2023 from S2ORC that contain an arXiv ID, resulting in $\sim$15k papers. S2ORC does not provide the publication date of the papers directly, so we use regex `2308' on the arXiv ID to extract papers posted in 08'23. We then use section annotations to get the section names and match using synonyms (`Related Work, Literature Review, Background') to extract section spans. We take the rest of the text as conditioning context except the related work section which results in $\sim$4.7k documents. Using the citation annotations, we extract the full text of cited papers from the S2ORC corpus again using corpus ID. Similar to Multi-XScience, we use paragraph annotations to create a dataset for the latest papers ($\sim$6.2k rows). We create a subset of 1,000 examples (RollingEval-Aug) where we have the content of all the cited papers. The average length of a related work summary is 95 words, while the average length of abstracts is 195. On average, we have 2 citations per example, which makes the dataset comparable to the original Multi-XScience dataset. 

\section{Other Generation Experiments}
\label{appendix:experiments}
\paragraph{Llama 2 fine-tuning}
In parallel, we also fine-tune Llama 2 models on the train set with the original shorter context, but they are very sensitive to hyperparameter configuration. When we instruct-finetune Llama 2 7B, it initially produces code. We find a slight improvement when fine-tuning the Llama 2 7B model for 30k steps with an LR of 5e-6 over 0-shot model (see Table~\ref{table:results-llama}), but it quickly overfits as we increase the LR or the number of steps. We leave hyperparameter optimization, fine-tuning larger models with RoPE scaling and plan-based generation for future work.

\begin{table}[ht]
\centering
\begin{tabular}{lccc}
\toprule 
\textbf{Model} & \textbf{ROUGE1 $\uparrow$} & \textbf{ROUGE2 $\uparrow$} & \textbf{ROUGEL $\uparrow$} \\
\midrule 
Llama 2-Chat 7B - 0-shot & 26.719 & 5.958 & 13.635 \\
Llama 2-Chat 7B - 10k steps (LR 5e-6) & 24.789 & 5.986 & 12.708 \\	
Llama 2-Chat 7B - 30k steps (LR 5e-6) & 27.795 & \textbf{6.601} & 14.409 \\	
\hdashline
Llama 2-Chat 7B - 60k steps (LR 1e-5) & 22.555 & 5.511 & 11.749 \\	
\bottomrule 
\end{tabular}
\caption{Results after fine-tuning Llama 2-Chat 7B on Multi-XScience dataset}
\label{table:results-llama}
\end{table}

\textbf{Longer context} While Llama 2 can ingest 4096 tokens, recent studies have found that it uses 19\% more tokens \citep{kadous_2023} than GPT-3.5 or GPT-4 (2048 and 4096 tokens respectively), implying that the effective number of words in Llama 2 is considerably lower than GPT-4 and only a bit higher than GPT-3.5. We experiment with the popular RoPE scaling \citep{su2021roformer} in 0-shot Llama models to increase the context length (4k--6k). This permits using the full text of the papers instead of just their abstracts. Results in Table~\ref{table:results-rope} show that directly using RoPE scaling on 0-shot models produces gibberish results. Instead, one needs to fine-tune the model with the longer context. In fact, a plan-based-longer-context CodeLlama (initialized from Llama 2 and trained with a 16k token context through RoPE scaling) improves on ROUGE1/L, but achieves comparable results as a shorter-context plan-based CodeLlama on ROUGE2. For reporting results with longer context Llama 2 using RoPE scaling ~\citep{su2021roformer}, we use HuggingFace Text Generation Inference.\footnote{\url{https://github.com/huggingface/text-generation-inference}}

\begin{table}[ht]
\centering
\begin{tabular}{lccc}
\toprule 
\textbf{Model} & \textbf{ROUGE1 $\uparrow$} & \textbf{ROUGE2 $\uparrow$} & \textbf{ROUGEL $\uparrow$} \\
\midrule 
Llama 2-Chat 7B (4000 words) & 17.844 & 1.835 & 10.149 \\	Llama 2-Chat 7B (5000 words) & 17.254 & 1.736 & 9.986 \\	Llama 2-Chat 7B (6000 words) & 17.179 & 1.647 & 9.897 \\	
\hdashline
Llama 2-Chat 13B (4000 words) & 20.071 & 3.516 & 10.916 \\	Llama 2-Chat 13B (5000 words) & 20.722 & 3.714 & 11.13 \\	Llama 2-Chat 13B (6000 words) & 17.179 & 1.647 & 9.897 \\	
\hdashline
Llama 2-Chat 70 (4000 words) & 19.916 & 2.741 & 10.456 \\	Llama 2-Chat 70B (5000 words) & 19.675 & 2.605 & 10.48 \\	Llama 2-Chat 70B (6000 words) & 20.437 & 2.976 & 10.756 \\
\hdashline
CodeLlama 34B-Instruct (4000 words) & 27.425 & 5.815 & 14.744 \\
\bottomrule 
\end{tabular}
\caption{Zero-shot results using RoPE scaling for larger context on RollingEval-Aug dataset. Here we report the max number of words used for truncation instead of the tokens.}
\label{table:results-rope}
\end{table}

\textbf{Code LLMs} We evaluate the performance of code-generating LLMs to write related-work sections requiring more formal and structured language. Since Code LLMs are pre-trained on text they might offer the best of both worlds. 
However, we observe that for our task, the models produce bibtex and Python code with relevant comments as part of the generated outputs. As shown in Table~\ref{table:results-code}, CodeLlama (34B Instruct) is good at following instructions and at generating natural language (ROUGE2 of 5.8 and 5.02 on Multi-XScience and RollingEval-Aug dataset). With a plan, CodeLlama even surpasses vanilla 0-shot Llama 2 70B (Table~\ref{table:results-rw-2308}).

\begin{table}[ht]
\centering
\resizebox{0.7\textwidth}{!}{
\begin{tabular}{lccc}
\toprule
\textbf{Model} & \textbf{ROUGE1 $\uparrow$} & \textbf{ROUGE2 $\uparrow$} & \textbf{ROUGEL $\uparrow$} \\
\midrule 
StarCoder & 12.485 & 1.104 & 6.532 \\
Lemur-70B & 15.172 & 2.136 & 7.411 \\
CodeLlama 34B-Instruct & 25.482 & 5.814 & 13.573 \\
\bottomrule 
\end{tabular}
}
\caption{0-shot results using code-based models on Multi-XScience dataset. CodeLlama performs reasonably well in generating natural language compared to the other code-based counterparts.}
\label{table:results-code}
\end{table}

\begin{figure}[htbp]
\centering 
\includegraphics[width=0.55\textwidth]{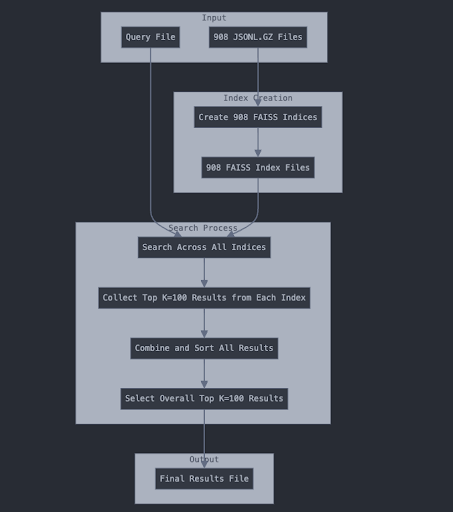}
\caption{Pipeline for creating FAISS indexes for 150M SPECTER2 embeddings.}
\label{fig:specter}
\end{figure}

\section{More implementation details}
\label{appendix:implementation}
\begin{figure}[htbp]
     \centering
     \includegraphics[width=0.65\textwidth]{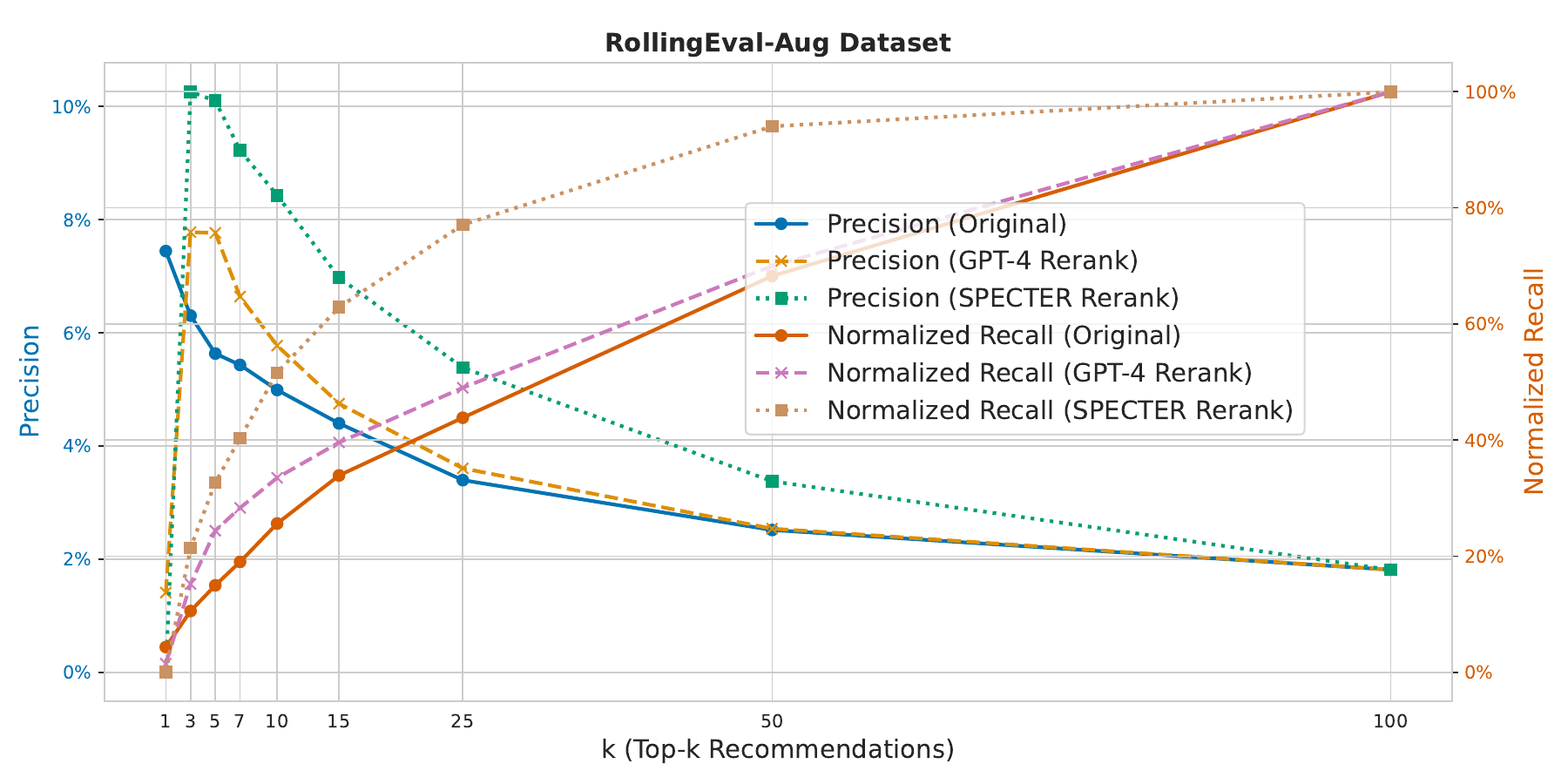}
    \caption{Effect of re-ranking strategies on the RollingEval-Aug dataset. We evaluate the Precision and Normalized Recall of the re-ranked results contrasting LLM-based based re-ranking with embedding-based ranker. We find a similar pattern as the RollingEval-Aug dataset.}
    \label{appendix-figure:retrieval-pr-curves}
\end{figure}

\subsection{Normalized Recall v/s Standard Recall: A Worked-out Example}
\label{app:recall-example}
Consider a query paper with the following statistics:
\begin{flalign*}
    & |\text{Ground Truth}| = n_{\text{gt}} = 84 \\  
    & |\text{Retrieved}| = 100 \\  
    & \text{Relevant Retrieved papers} = |\text{Retrieved} \cap \text{Ground Truth}| = c = 10 \\  
    & \text{Relevant papers in top-40} = n_{\text{rel}} = 4  
\end{flalign*}

Using these values, we compute the metrics at \( k=40 \):
\begin{flalign*}
    & \text{Precision@40} = \frac{n_{\text{rel}}}{40} = \frac{4}{40} = 0.010; \quad \text{Normalized Recall@40} = \frac{n_{\text{rel}}}{c} = \frac{1}{10} = 0.100; \quad \text{Recall} = \frac{n_{rel}}{n_{gt}} = \frac{4}{84} = 0.048
\end{flalign*}

This example illustrates how Normalized Recall@k differs from standard recall. Instead of being limited by the total number of ground truth citations, it evaluates how well the method ranks the retrievable relevant papers. In this case, despite a low precision, the normalized recall is relatively high, indicating that the method effectively ranks the relevant papers it does retrieve.

\subsection{Generation Implementation}
We use HuggingFace Transformers and PyTorch~
~\citep{paszke2017automatic} for our experiments.\footnote{Code will be released at \url{github.com}} 
We calculate ROUGE scores~\citep{lin-2004-rouge} using the Huggingface~\citep{wolf2019huggingface} evaluate library\footnote{\url{https://huggingface.co/spaces/evaluate-metric/rouge} Since it is a known issue in the NLG community of different implementations producing different results, we stick to evaluate==0.4.0 for reporting all the results, reproducing the ROUGE scores for baselines from Multi-XScience model outputs.}. To split sentences, we use `en\_core\_web\_sm' model from SpaCy\footnote{\url{https://spacy.io/usage/linguistic-features}}. Additionally, we use Anyscale endpoints\footnote{\url{https://app.endpoints.anyscale.com/}} to generate 0-shot Llama 2 results and OpenAI API\footnote{\url{https://platform.openai.com/docs/guides/gpt}} to generate results for GPT-3.5-turbo and GPT-4. 

\subsection{Demo implementation}
We build our system using the ReactJS framework, which provides a nice interface to build system demos quickly and efficiently. More details about the demo implementation can be found in our system paper~\cite{agarwal2024litllm} and the project page. 

We query the Semantic Scholar API available to search for the relevant papers. Specifically, we use the Academic Graph\footnote{\url{https://api.semanticscholar.org/api-docs/graph}} and Recommendations\footnote{\url{https://api.semanticscholar.org/api-docs/recommendations}} API endpoint. We use OpenAI API to generate results for LLM using GPT-3.5-turbo and GPT-4 models. At the same time, our modular pipeline allows using any LLM (proprietary or open-sourced) for different components. We also allow the end-user to sort the retrieved papers by relevance (default S2 results), citation count, or year. More details about the demo system can be found in our system paper.

\subsection{SPECTER Implementation}
\label{appendix:specter}
We build an index of 150M SPECTER2 embeddings that we can use as an alternative to both a search engine and a prompting-based ranking module.
Figure~\ref{fig:specter} shows our pipeline for creating the index.
Specifically, the SPECTER2 database comes with 908 json.gz files containing compressed embeddings.
For each json.gz file, we construct a FAISS index that we can query for the nearest neighbors of a given query embedding.
We perform index construction in a multi-threaded manner to speed up the process.
Upon constructing a FAISS index for all the json.gz files, we iterate over each query paper, search for the top 100 relevant papers using the SPECTER embeddings in \emph{each} FAISS index, and then finally merge the results to get the top 1000 papers for each query paper.

\subsection{Comparative analysis of the computational costs}
\label{appendix:costs}

We compare the costs of different LLMs for both stages in Table \ref{table:cost}. 

\textbf{Ranking:} We explore two types of LLM-based reranking mechanisms: permutation and debate ranking.
For $n$ query papers (=500 for our RollingEval datasets) and top-$k$ candidates retrieved from S2 per query paper ($k$=100 in our experiments), permutation ranking would require $n$ API calls, whereas debate ranking would require $n*k$ API calls.
Debate ranking needs more API calls as it involves one additional API call per candidate paper to generate the citation probability score and reasoning.
Therefore, there are $k$ additional API calls per query paper compared to permutation ranking, where we prompt the LLM to directly rank relevance for all the candidate papers.
We refer the reader to Figure~\ref{appendix-figure:debate_prompt}for The exact prompt used for debate ranking.

\textbf{Generation:}
There was only one request per query abstract in the RollingEval dataset, so 500 requests in total for each experiment (as $n=500$ in RollingEval). The table below summarizes the API analysis for the two stages of the pipeline for the RollingEval experiments.

\begin{table*}[ht]
\centering
\resizebox{0.85\textwidth}{!}{
\begin{tabular}{llrcr}
\toprule
\textbf{Experiment} & \textbf{Method} & \textbf{Requests} & \textbf{Tokens} & \textbf{Cost} \\
\midrule 
\multirow{3}{*}{Ranking}
& GPT-4 Permutation Reranking & 500 & $\sim$20M input + $\sim$0.25M output tokens & \$50 \\
& Llama-3.1 Debate Ranking (w/o attribution) & 500 x 100 & $\sim$33M input + $\sim$0.25M output tokens & \$0 \\
& Llama-3.1 Debate Ranking (w/ attribution) & 500 x 100 & $\sim$33M input + $\sim$15M output tokens & \$0 \\
\midrule 
\multirow{4}{*}{Generation}
& Llama 2 70B (using Anyscale Endpoint) & 500 & $\sim$0.75M input + $\sim$0.15M output tokens & \$3.84 \\
& GPT-3.5-turbo & 500 & $\sim$0.75M input + $\sim$0.15M output tokens & \$4.2 \\
& GPT-4 & 500 & $\sim$0.75M input + $\sim$0.15M output tokens & \$22 \\
& GPT-4 (Plan) & 500 & $\sim$0.75M input + $\sim$0.15M output tokens & \$25 \\
\bottomrule
\end{tabular}}
\caption{Computational costs for different experiments on the RollingEval dataset. Costs for generation experiments on the Multi-XScience are approximately 10 times that of the RollingEval dataset.}
\label{table:cost}
\end{table*}

\begin{table}[ht]
\centering
\resizebox{\textwidth}{!}{ 
\footnotesize{ 
 \begin{tabular}{p{0.9\textwidth}}
  \hline
  \textbf{Abstract of Multi-XScience paper \citep{lu-etal-2020-multi-xscience}}\\\hline 		 
\textbf{\textcolor{orange}{Reference @cite\_1:}} Multi-document summarization is a challenging task for which there exists little large-scale datasets. We propose Multi-XScience, a large-scale multi-document summarization dataset created from scientific articles. MultiXScience introduces a challenging multi-document summarization task: writing the related-work section of a paper based on its abstract and the articles it references. Our work is inspired by extreme summarization, a dataset construction protocol that favours abstractive modeling approaches. Descriptive statistics and empirical results—using several state-of-the-art models trained on the MultiXScience dataset—reveal that Multi-XScience is well suited for abstractive models.
\\\hline
  \textbf{Abstract of Extractive and Abstractive Summarization paper \citep{pilault-etal-2020-extractive}}\\\hline 		 
\textbf{\textcolor{blue}{Reference @cite\_2:}} We present a method to produce abstractive summaries of long documents that exceed several thousand words via neural abstractive summarization. We perform a simple extractive step before generating a summary, which is then used to condition the transformer language model on relevant information before being tasked with generating a summary. We show that this extractive step significantly improves summarization results. We also show that this approach produces more abstractive summaries compared to prior work that employs a copy mechanism while still achieving higher rouge scores. Note: The abstract above was not written by the authors, it was generated by one of the models presented in this paper.
\\\hline
  \textbf{Abstract of Galactica paper \citep{taylor2022galactica}}\\\hline 		 
\textbf{\textcolor{purple}{Reference @cite\_3:}} Information overload is a major obstacle to scientific progress. The explosive growth in scientific literature and data has made it ever harder to discover useful insights in a large mass of information. Today scientific knowledge is accessed through search engines, but they are unable to organize scientific knowledge alone. In this paper we introduce Galactica: a large language model that can store, combine and reason about scientific knowledge. We train on a large scientific corpus of papers, reference material, knowledge bases and many other sources. We outperform existing models on a range of scientific tasks. On technical knowledge probes such as LaTeX equations, Galactica outperforms the latest GPT-3 by 68.2\% versus 49.0\%. Galactica also performs well on reasoning, outperforming Chinchilla on mathematical MMLU by 41.3\% to 35.7\%, and PaLM 540B on MATH with a score of 20.4\% versus 8.8\%. It also sets a new state-of-the-art on downstream tasks such as PubMedQA and MedMCQA dev of 77.6\% and 52.9\%. And despite not being trained on a general corpus, Galactica outperforms BLOOM and OPT-175B on BIG-bench. We believe these results demonstrate the potential for language models as a new interface for science. We open source the model for the benefit of the scientific community.
\\\hline
  \textbf{Plan for Table \ref{table:qualitative}}\\\hline 		 Please generate 5 sentences in 120 words. Cite \textcolor{orange}{@cite\_1} at line 1, 3 and 5.
\\\hline
  \textbf{Plan for Table \ref{table:qualitative}}\\\hline 		 Please generate 5 sentences in 120 words. Cite \textcolor{orange}{@cite\_1} at line 1 and 3. Cite \textcolor{blue}{@cite\_2} at line 2 and 5. Cite \textcolor{purple}{@cite\_3} at line 4 and 5.
\\\hline
  \end{tabular}
  }}
\caption{Abstracts of papers which are reference citations in Tables \ref{table:qualitative} and \ref{table:qualitative-cite3}.}
\label{table:qualitative-ref-abstracts}
\end{table}

\begin{table}[ht]
\centering
\resizebox{\textwidth}{!}{ 
\scriptsize{ 
 \begin{tabular}{p{0.9\textwidth}}
  \hline
\textbf{GPT-4 0-shot} \\ \hline
The field of multi-document summarization has been a challenging one due to the lack of large-scale datasets (\textcolor{orange}{@cite\_1}). One significant attempt to address this gap is the Multi-XScience dataset, a large-scale collection of scientific articles designed to simulate the task of writing the related-work section of a research paper (\textcolor{orange}{@cite\_1}). This approach was inspired by extreme summarization, a methodology that promotes abstractive modeling approaches. However, the current study offers a novel strategy for literature review generation using large language models (LLMs), which goes beyond the scope of the Multi-XScience dataset. The proposed method involves the creation of an intermediate plan, which is then used to generate the actual text. Empirical evidence suggests that this approach improves the quality of the generated literature reviews compared to the conventional zero-shot generation. Additionally, this work expands the Multi-XScience corpus to include the full text of research papers, further enhancing the depth and breadth of information available for literature review generation. This study also ensures the validity of zero-shot experiments by creating a new test corpus from recent arXiv papers, thus avoiding test set contamination. This reflects a more comprehensive and rigorous approach to the generation of literature reviews using LLMs.
\\\hline
\textbf{GPT-3.5 Plan} \\ \hline
The task of generating literature reviews using large language models (LLMs) has gained attention recently. One approach is to condition the LLM on the abstracts of the papers to be cited and generate the literature review directly. However, the authors propose a novel strategy where they first generate a plan for the review and then use it to generate the actual text. This intermediate planning step is found to improve the quality of the generated literature reviews compared to vanilla zero-shot generation. The authors extend the Multi-XScience corpus, previously containing only abstracts, to include the full text of research papers. Additionally, they create a new test corpus consisting of recent arXiv papers, ensuring that their zero-shot experiments are not affected by contamination from the test set. The Multi-XScience dataset, used in their work, is particularly suitable for abstractive models due to its challenging multi-document summarization task (\textcolor{orange}{@cite\_1}). The empirical results from several state-of-the-art models trained on this dataset demonstrate its effectiveness for abstractive modeling approaches.		 
\\\hline
\textbf{GPT-3.5 0-shot} \\ \hline
Prior work in the field of generating literature reviews has primarily focused on the use of large language models (LLMs) and multi-document summarization.  Multi-document summarization is a challenging task due to the lack of large-scale datasets (\textcolor{orange}{@cite\_1}). The Multi-XScience dataset was proposed as a solution to this problem, specifically designed for generating the related-work section of a paper based on its abstract and the articles it references. This dataset introduced a challenging multidocument summarization task that favors abstractive modeling approaches (\textcolor{orange}{@cite\_1}).

Our work builds upon these prior efforts by exploring the zero-shot abilities of recent LLMs for the task of generating literature reviews. We propose a novel strategy that involves generating an intermediate plan for the review, which is then used to generate the actual text. While modern LLMs can generate literature reviews without such intermediate plans, our empirical study shows that the use of these plans improves the quality of the generated reviews compared to vanilla zero-shot generation.

In addition to extending the Multi-XScience corpus to include the full text of research papers, we also create a new test corpus consisting of recent arXiv papers. This allows us to ensure that our zero-shot experiments are not affected by test set contamination.

The prior work on multi-document summarization and the Multi-XScience dataset provides a valuable foundation for our study. However, they do not specifically address the use of intermediate plans for literature review generation, which is a key contribution of our work. By incorporating these intermediate plans, we aim to enhance the quality and coherence of the generated literature reviews.
\\\hline
  \textbf{Llama 2 Plan} \\ \hline
The task of generating literature reviews for scientific research papers has gained significant attention in recent years, with a focus on using large language models (LLMs) for zero-shot generation. Inspired by extreme summarization, Multi-XScience was introduced as a large-scale dataset for multi-document summarization, including the challenging task of writing related-work sections based on abstracts and referenced articles. Our work builds upon this foundation by exploring the effectiveness of intermediate plans for literature review generation. Our empirical study shows that these plans improve the quality of generated literature reviews, and we contribute to the extension of the Multi-XScience corpus to include full-text papers. Additionally, we create a new test corpus to ensure zero-shot experiments are free from test set contamination. (\textcolor{orange}{@cite\_1})
  \\ \hline
\textbf{Llama 2 0-shot}\\ \hline
In recent years, there has been growing interest in using large language models (LLMs) for zero-shot literature review generation. However, most studies have relied on training or prompting LLMs to condition on all abstracts of papers to be cited, which can be limiting in terms of quality and efficiency. In contrast, our proposed approach involves generating a plan for the literature review and then using it to generate the actual text. We evaluate the effectiveness of this approach using a novel extended version of the Multi-XScience corpus, which now includes the full text of research papers. Our results show that intermediate plans improve the quality of generated literature reviews over vanilla zero-shot generation.
One of the key challenges in multi-document summarization is the lack of large-scale datasets. To address this, we have created a new test corpus consisting of recent arXiv papers with full content, ensuring that our zero-shot experiments are not contaminated by the release of the LLMs used in our study.
Our work builds on the concept of extreme summarization, which favors abstractive modeling approaches. We have employed several state-of-the-art models trained on the Multi-XScience dataset to evaluate the effectiveness of our approach. The results show that our dataset is well-suited for abstractive models, providing a valuable resource for researchers working in this area. 
In summary, our contribution includes a novel approach to literature review generation using LLMs, an extended version of the Multi-XScience corpus, and a new test corpus. Our results demonstrate the effectiveness of our approach and highlight the potential of using LLMs for zero-shot literature review generation.
References:
* \textcolor{orange}{@cite\_1}: Multi-XScience: A Large-Scale Multi-Document Summarization Dataset.
* \textcolor{red}{@cite\_2: Extreme Summarization: A Dataset Construction Protocol for Abstractive Modeling.}
  
  \\\hline
  \end{tabular}
  }}
\caption{We show outputs of different models corresponding to the same plan and prompt as paragraph of our related work (by GPT-4 Plan). Llama 2 70B 0-shot has the tendency to hallucinate citation (in red) which is remedied in plan based generation. Abstract of reference paper shown in Table \ref{table:qualitative-ref-abstracts} while the main abstract is a slight modified version of our current abstract. We also find that the 0-shot results are more verbose, copying the exact text verbatim, while the generated literature review using plans are more representative of the way humans write literature review.}
\label{table:qualitative}
\end{table}

\begin{table}[ht]
\centering
\resizebox{\textwidth}{!}{ 
\scriptsize{ 
 \begin{tabular}{p{0.92\textwidth}}
  \hline
  \textbf{GPT-4 Plan}\\\hline 		 
  The concept of generating the related-work section from abstracts and referenced articles is not entirely new, as evidenced by the Multi-XScience dataset proposed by \textcolor{orange}{@cite\_1}. The approach of extending abstractive summarization to lengthy documents via an extractive step for conditioning, as described by \textcolor{blue}{@cite\_2}, aligns with the proposed methodology here. However, the strategy introduced in this study differentiates itself by creating an iterative plan prior to the generation of the review, a step not present in the work of \textcolor{orange}{@cite\_1}. The LLM used in this study also builds upon the capabilities of models like Galactica \textcolor{purple}{@cite\_3}, which have demonstrated exceptional performance on scientific tasks. The results of \textcolor{blue}{@cite\_2} and \textcolor{purple}{@cite\_3} support the potential of large language models in scientific literature review generation, while the proposed approach aims to further enhance the quality of these outputs through structured planning.
\\ \hline 
\textbf{GPT-4 0-shot} \\ \hline
The literature review generation task, a form of multi-document summarization, has been tackled through different strategies. \textcolor{orange}{@cite\_1} proposed a dataset, MultiXScience, to foster abstractive approaches, while \textcolor{blue}{@cite\_2} used an extractive step to enhance neural summarization. Our work complements these by introducing intermediate plans to improve generation quality. Meanwhile, \textcolor{purple}{@cite\_3}'s Galactica demonstrated potential for language models in scientific tasks, a direction we explore with zero-shot experiments.
\\\hline
\textbf{GPT-3.5 Plan} \\ \hline
The task of generating a literature review using large language models (LLMs) has gained attention recently. One approach is to condition the LLM on the abstract and content of related papers to generate the review (\textcolor{orange}{@cite\_1}). However, it has been observed that generating an intermediate plan for the review improves its quality compared to vanilla zero-shot generation. This motivated the proposal and examination of a novel strategy where a plan is first generated and then used to generate the actual text (\textcolor{orange}{@cite\_1}). Another method to improve abstractive summarization is by performing an extractive step before generating the summary. This step conditions the transformer language model on relevant information, leading to better summarization results compared to prior work that employs a copy mechanism (\textcolor{blue}{@cite\_2}). Additionally, the use of large language models like Galactica has shown promising results in organizing and reasoning about scientific knowledge (\textcolor{purple}{@cite\_3}). Galactica outperforms existing models on technical knowledge probes and downstream tasks, demonstrating the potential for language models as a new interface for science (\textcolor{purple}{@cite\_3}).
\\\hline
\textbf{GPT-3.5 0-shot} \\ \hline
Prior work has explored the use of large language models (LLMs) for various natural language processing tasks. Multi-document summarization is one such task, and the Multi-XScience dataset (\textcolor{orange}{@cite\_1}) has been created specifically for this purpose. This dataset focuses on writing the related-work section of a scientific paper based on its abstract and the articles it references. The dataset construction protocol favors abstractive modeling approaches, making it suitable for LLMs. Another approach to abstractive summarization is presented in a paper (\textcolor{blue}{@cite\_2}), where the authors propose a method to generate summaries of long documents. They employ an extractive step to gather relevant information before conditioning the transformer language model for summary generation. This approach yields more abstractive summaries compared to prior work, while still achieving higher rouge scores. Furthermore, the concept of using LLMs for scientific knowledge organization and reasoning is introduced in the Galactica model (\textcolor{purple}{@cite\_3}). This model is trained on a large scientific corpus and outperforms existing models on various scientific tasks. It demonstrates superior performance on technical knowledge probes and reasoning tasks, surpassing GPT-3 and other models. Galactica also achieves state-of-the-art results on downstream tasks such as PubMedQA and MedMCQA, showcasing the potential of LLMs as a new interface for science. Based on these prior works, we propose a novel strategy for literature review generation using LLMs. We first generate a plan for the review and then use it to generate the actual text. While modern LLMs can generate literature reviews without intermediate plans, our empirical study indicates that these plans improve the quality of generated reviews. Additionally, we create a new test corpus consisting of recent arXiv papers, ensuring that our zero-shot experiments are not affected by test set contamination. 
\\\hline
\textbf{Llama 2 Plan} \\ \hline
In recent years, there has been a growing interest in using large language models (LLMs) for scientific literature review generation (\textcolor{orange}{@cite\_1}). One approach is to use a novel strategy that involves generating a plan for the review and then using it to generate the actual text (\textcolor{orange}{@cite\_1}). This approach has been shown to improve the quality of generated literature reviews over vanilla zero-shot generation (\textcolor{orange}{@cite\_1}). Another approach is to use a simple extractive step before generating a summary, which is then used to condition the transformer language model on relevant information (\textcolor{blue}{@cite\_2}). This approach has been shown to produce more abstractive summaries compared to prior work that employs a copy mechanism while still achieving higher ROUGE scores (\textcolor{blue}{@cite\_2}). Additionally, there have been efforts to train LLMs on large scientific corpora to improve their ability to store, combine, and reason about scientific knowledge (\textcolor{purple}{@cite\_3}). For example, Galactica, a large language model, has been trained on a range of scientific tasks and has outperformed existing models on technical knowledge probes, reasoning, and downstream tasks (\textcolor{purple}{@cite\_3}). These advancements in LLMs have the potential to revolutionize the way we interact with scientific literature and accelerate scientific progress.
\\\hline
\textbf{Llama 2 0-shot} \\ \hline
In this section, we discuss the related work in the field of scientific literature review generation using large language models (LLMs). Recent works have explored the use of LLMs for multi-document summarization tasks, such as generating the related work section of a paper based on its abstract and the articles it references (\textcolor{orange}{@cite\_1}). These works have shown that LLMs can produce high-quality summaries, but may suffer from the lack of relevant information and context (\textcolor{blue}{@cite\_2}). To address this issue, we propose a novel strategy for literature review generation that involves generating a plan for the review before generating the actual text. \textcolor{red}{This approach has been shown to improve the quality of generated literature reviews over vanilla zero-shot generation }(\textcolor{purple}{@cite\_3}). Additionally, we create a new test corpus consisting of recent arXiv papers to ensure that our zero-shot experiments do not suffer from test set contamination. Our work builds upon the recent advances in LLMs and their ability to store, combine, and reason about scientific knowledge (\textcolor{purple}{@cite\_3}). We believe that our approach has the potential to revolutionize the way scientific literature is reviewed and organized, and we open-source our model for the benefit of the scientific community. In summary, our related work demonstrates the potential of LLMs for scientific literature review generation and highlights the importance of context and information in producing high-quality summaries. Our proposed approach aims to address these limitations and provide a more effective way of generating literature reviews using LLMs.
\\\hline
  \end{tabular}
  }}
\caption{We show outputs of different models corresponding to the reference cited abstracts and plan mentioned in Table \ref{table:qualitative-ref-abstracts} while the main abstract is a slightly modified version of our current abstract. In this example, though, we have all the citations covered by all the models, we can see GPT-4 (Plan) output to be concise and closely following the plan. Moreover, GPT-3.5 and Llama 0-shot outputs are excessively elaborate, making them unsuitable for inclusion in the literature review. Our findings indicate that while LLMs can help in certain aspects, the final output still heavily relies on inputs from a human researcher.
}
\label{table:qualitative-cite3}
\end{table}

\begin{figure}[h]
\centering
\includegraphics[width=\linewidth]{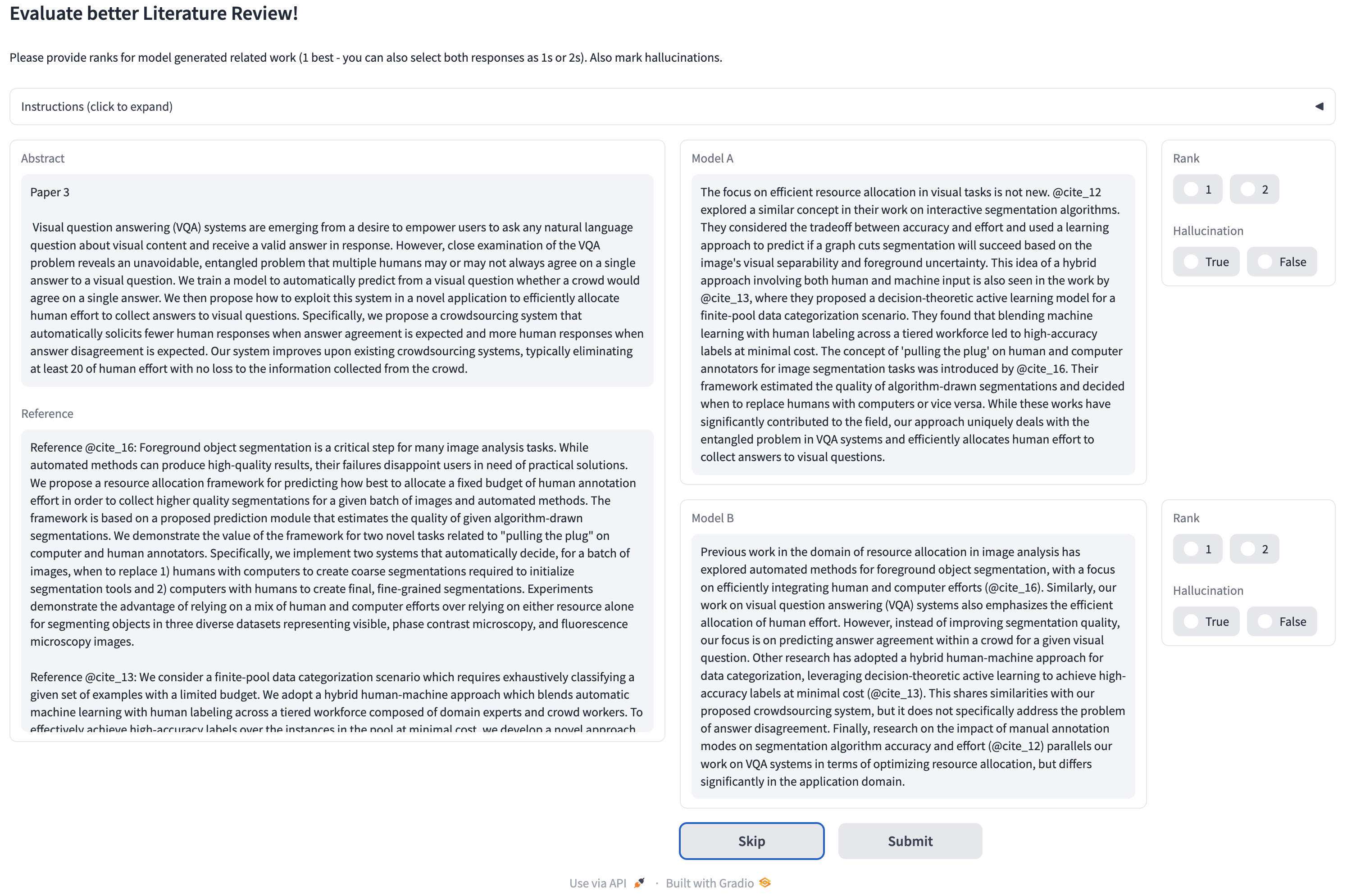}
\caption{Interface of our human evaluation setup.}
\label{fig:interface-human-eval}
\end{figure} 

\begin{figure}[h]
\centering
\includegraphics[width=0.8\linewidth]{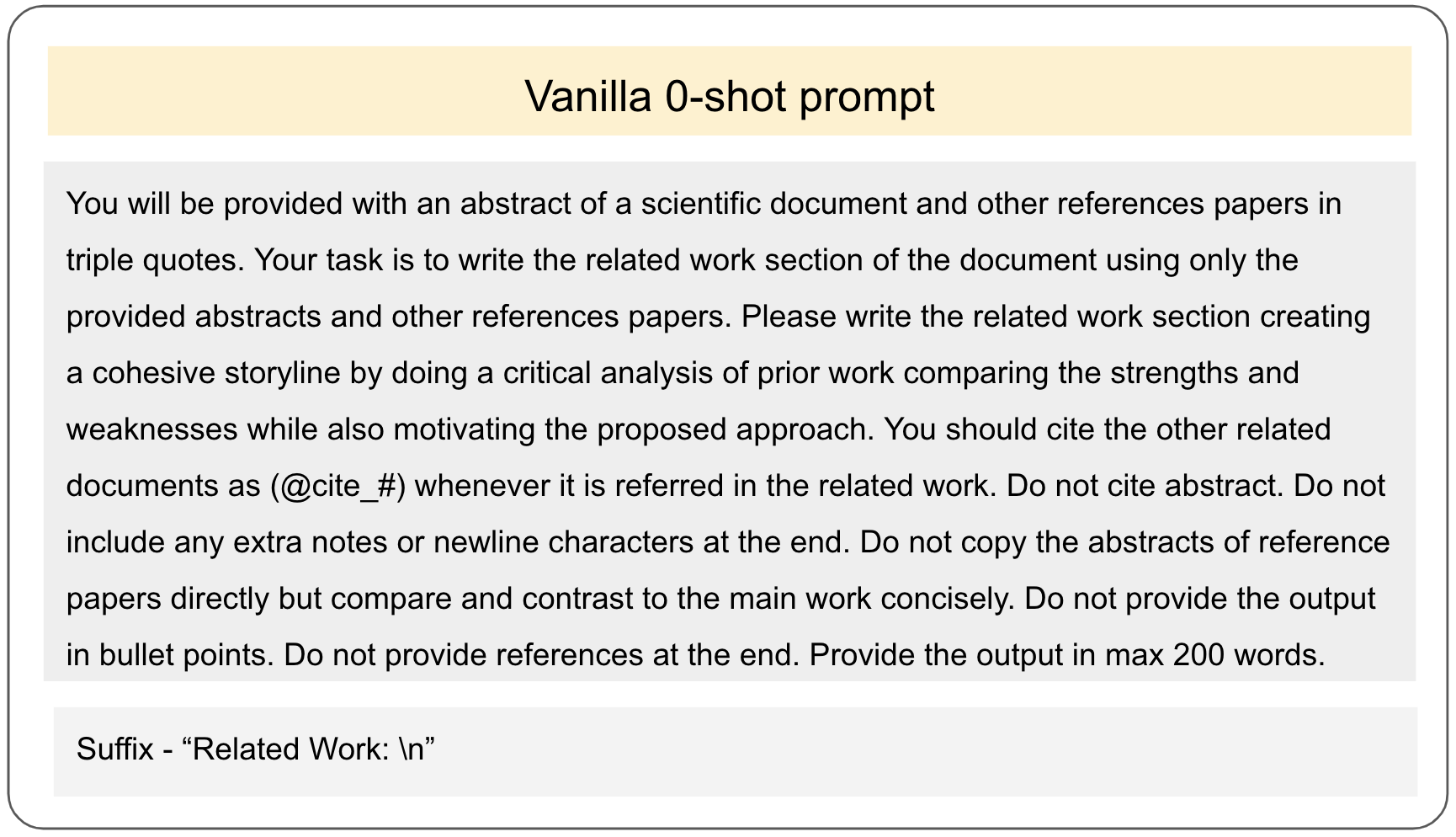}
\caption{Prompt used for Vanilla 0-shot generation.}
\label{fig:prompt-0-shot}
\end{figure}

\begin{figure}[h]
\centering
\includegraphics[width=0.95\linewidth]{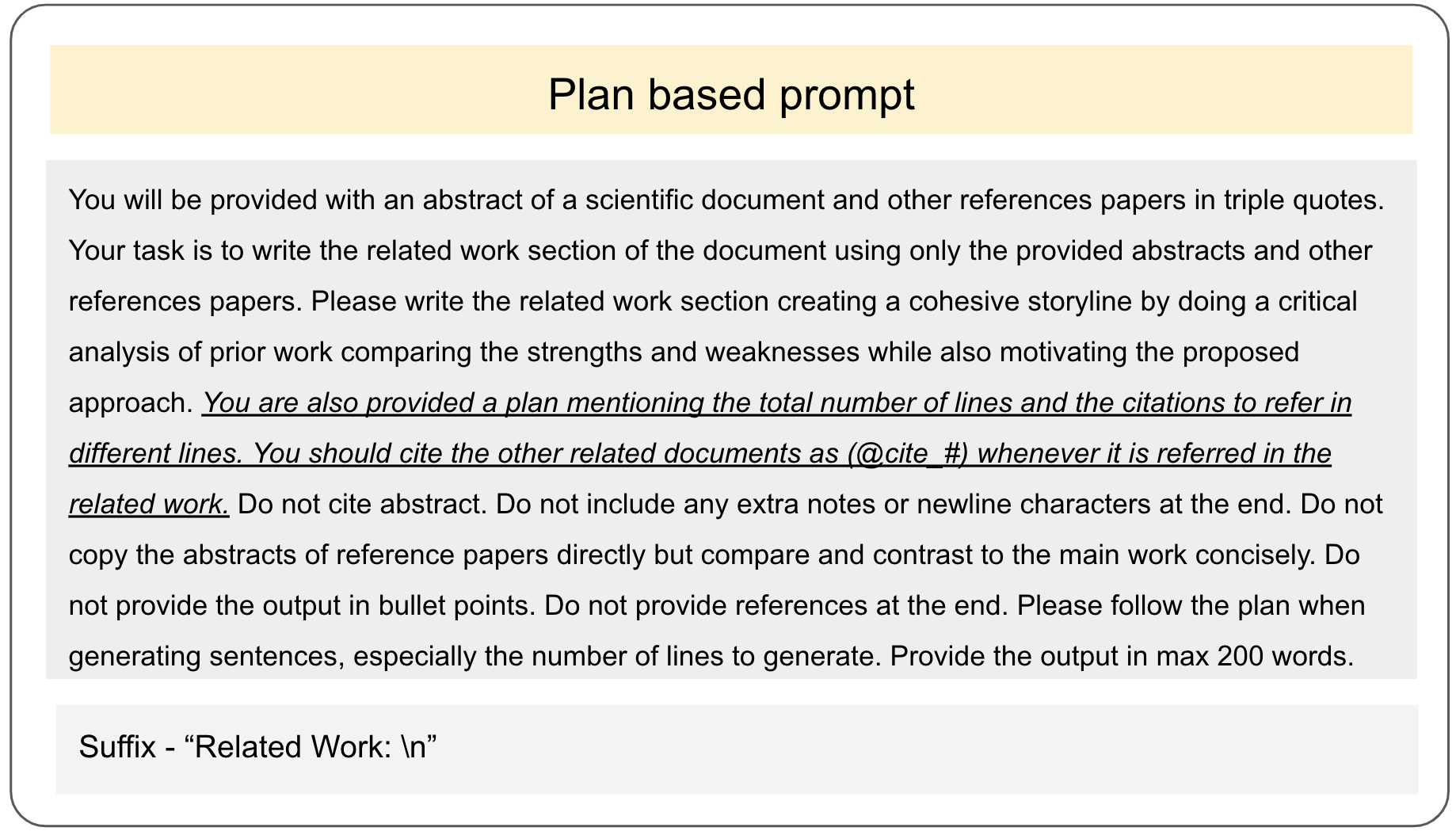}
\caption{Prompt used for plan-based generation. Underlined text shows the variation compared to the vanilla 0-shot prompting, where the user provides a structure of the expected paragraph.}
\label{fig:prompt-plan}
\end{figure}

\begin{figure}[h]
\centering
\includegraphics[width=0.95\linewidth]{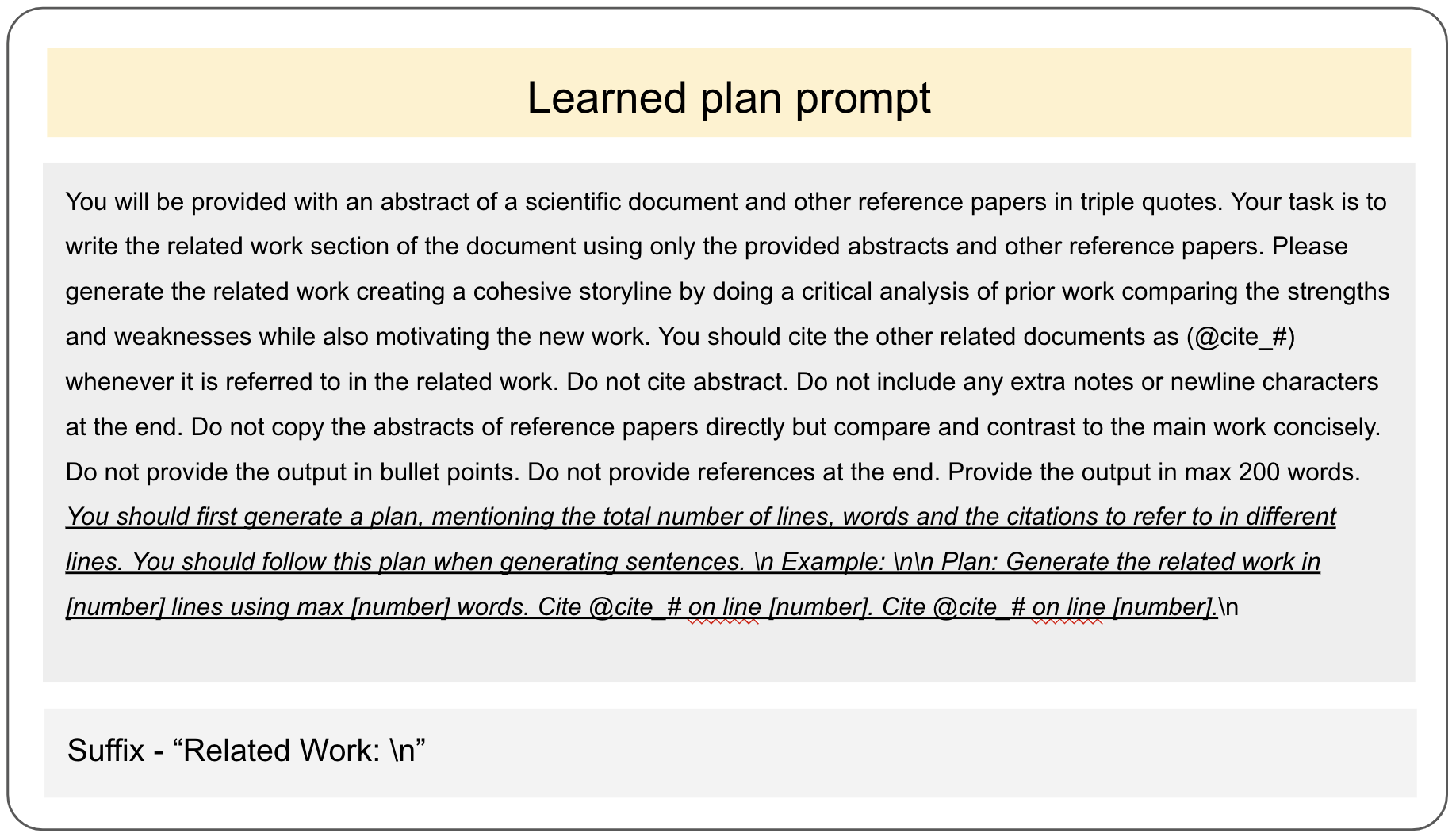}
\caption{Prompt used when the plan is learned during generation. The model first generates a plan of sentences and citations which it would then condition upon to generate the final related work text, which can be considered as an extension of CoT style thinking step by step.}
\label{fig:prompt-learned-plan}
\end{figure} 

\begin{figure}[h]
\centering
\includegraphics[width=0.95\linewidth]{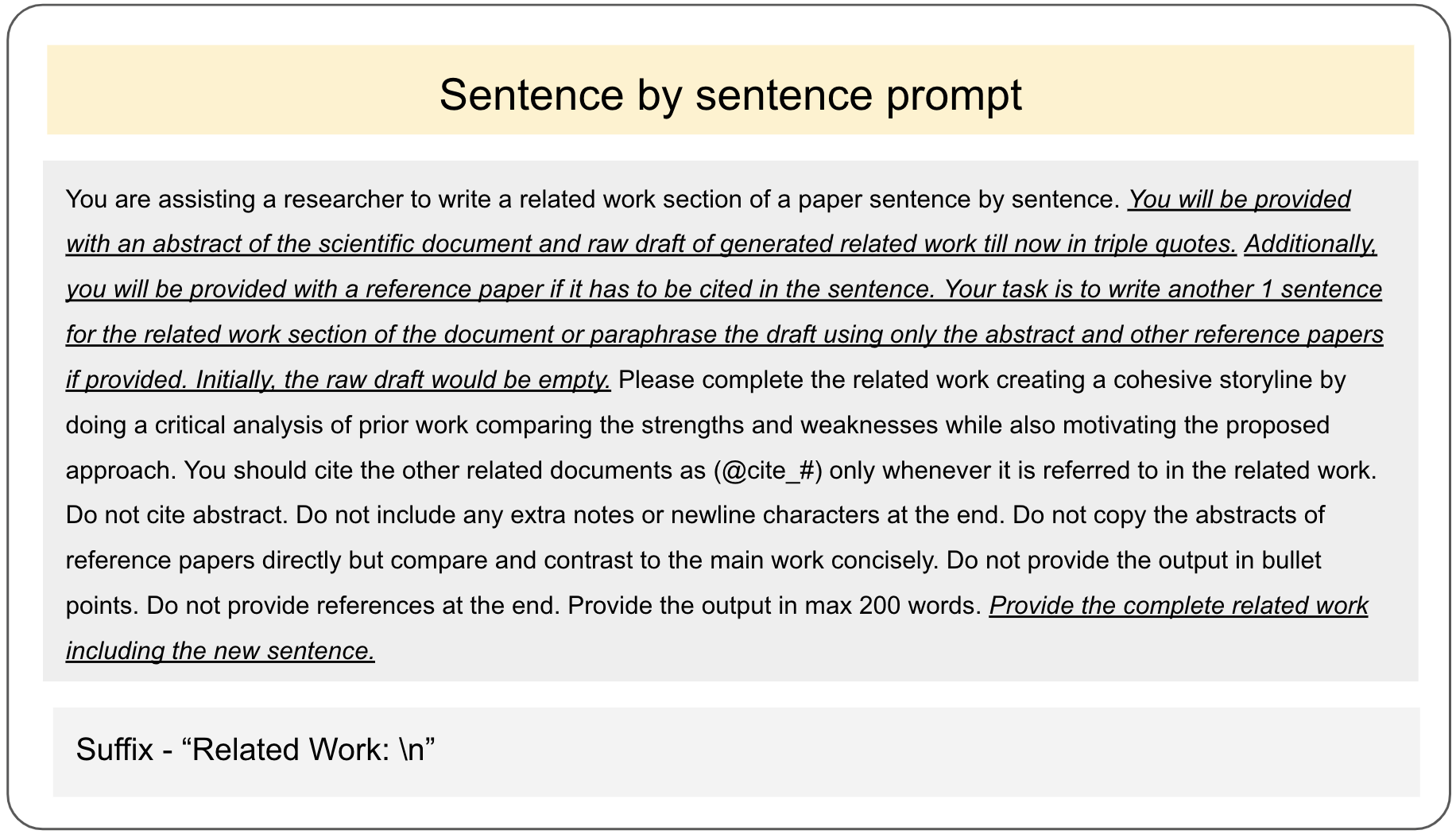}
\caption{Prompt used for sentence-by-sentence generation. In this scenario, we prompt the model to generate one sentence for each citation individually.}
\label{fig:prompt-sentence-sentence}
\end{figure} 

\begin{figure}[h]
\centering
\includegraphics[width=0.95\linewidth]{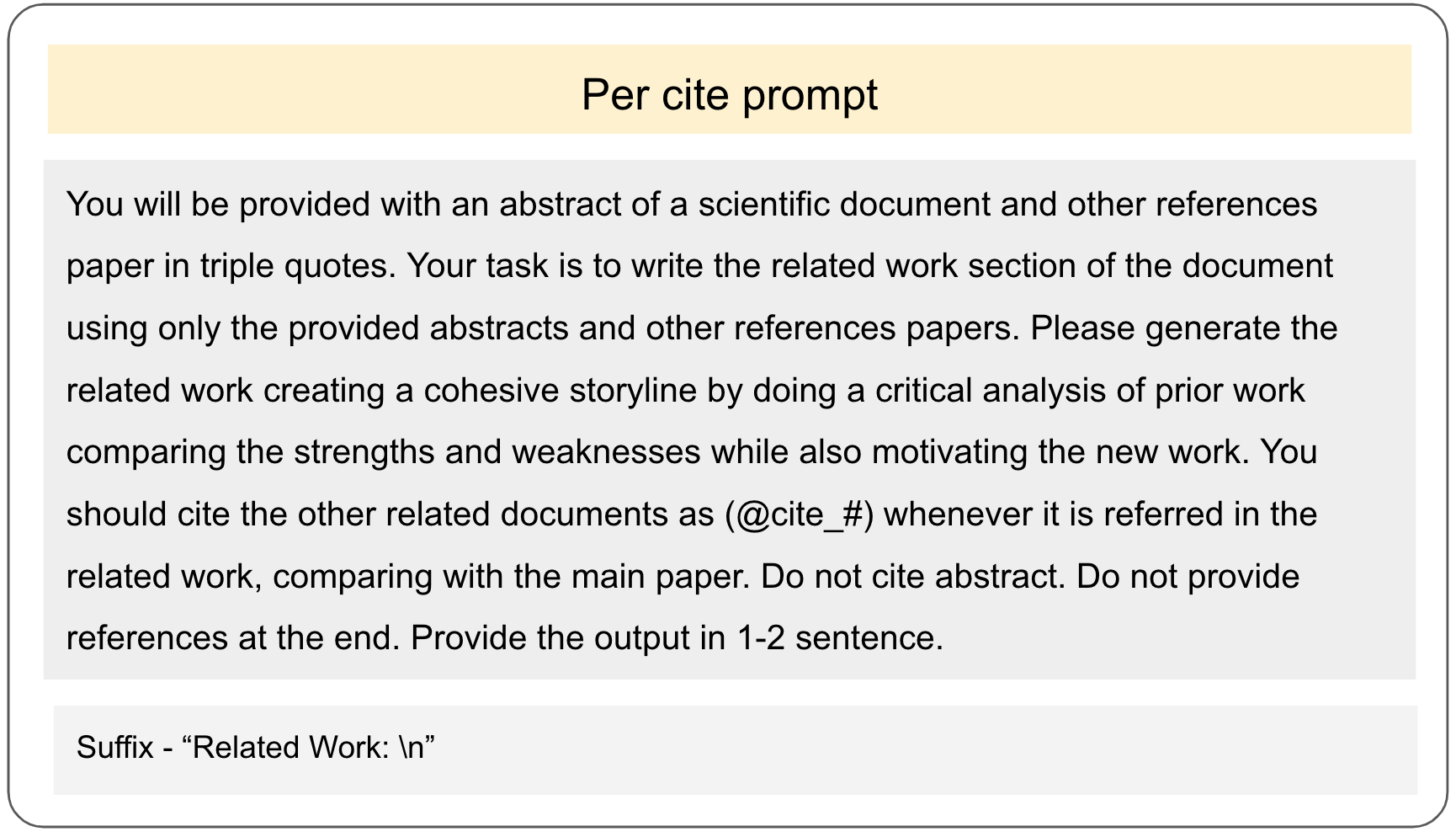}
\caption{Prompt used for generating output per citation.}
\label{fig:prompt-per-cite}
\end{figure} 

\begin{figure}[ht]
\centering
\includegraphics[width=0.8\linewidth]{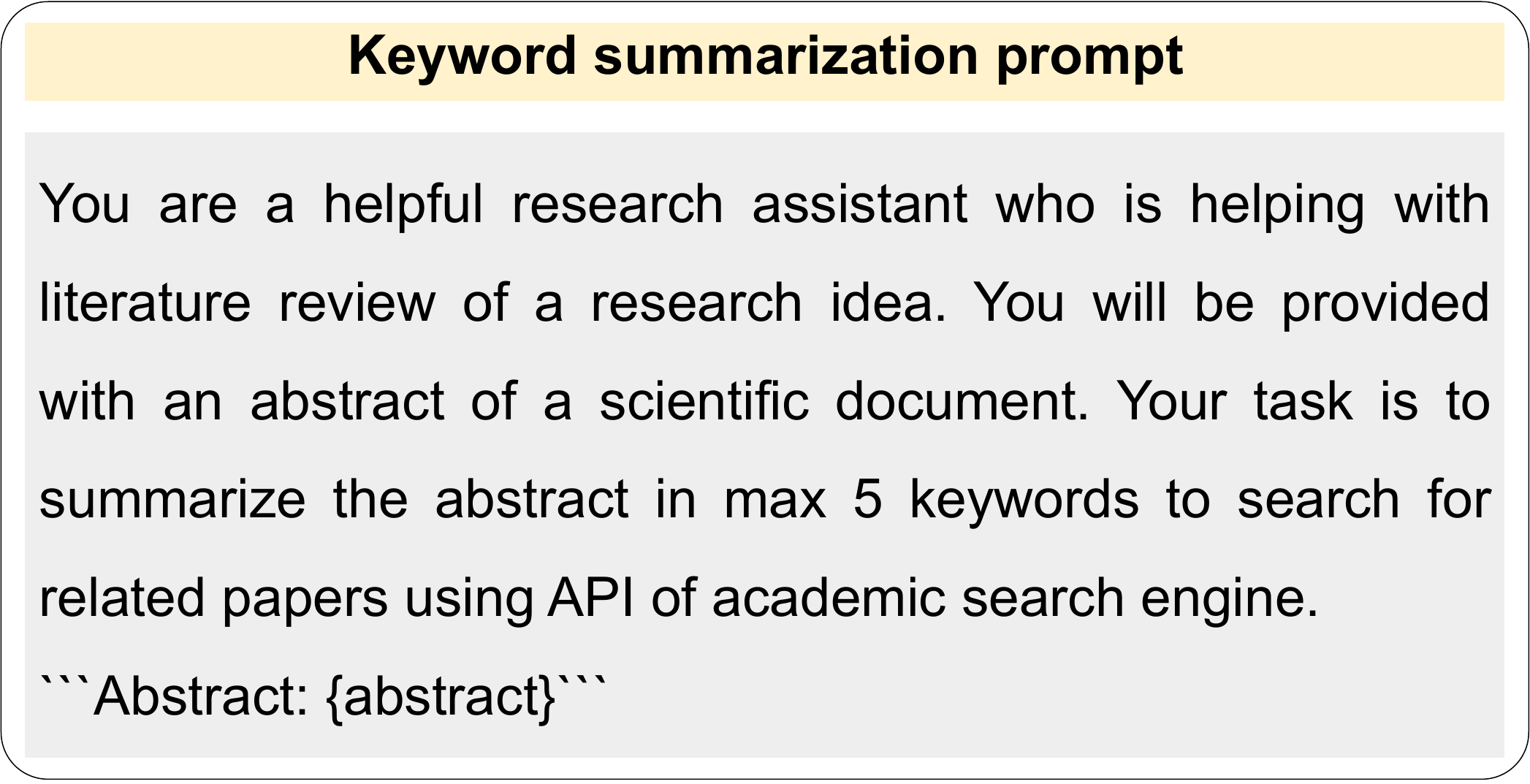}
\caption{Prompt used to summarize the research idea by LLM to search an academic engine}
\label{fig:summarization-prompt}
\end{figure} 

\begin{figure*}
\begin{lstlisting}[breaklines=true]
"""
You are a helpful research assistant who is helping with literature review of a research idea. Your task is to rank some papers based on their relevance to the query abstract.


## Instruction:
Given the query abstract:
<query_abstract>{query_abstract}</query_abstract>

Given the candidate reference paper abstract:
<candidate_paper_abstracts>{reference_papers}</candidate_paper_abstracts>

* Given the abstract of the candidate reference papers, provide me with a number between 0 and 100 (upto two decimal places) that is proportional to the probability of a paper with the given query abstract including the candidate reference paper in its literature review.
* In addition to the probability, give me arguments for and against including this paper in the literature review.
* You must enclose your arguments for including the paper within <arguments_for> and </arguments_for> tags.
* You must enclose your arguments for including the paper within <arguments_against> and </arguments_against> tags.
* Extract relevant sentences from the candidate paper abstract to support your arguments.
* Put the extracted sentences in quotes.
* You can use the information in other candidate papers when generating the arguments for a candidate paper.
* You must enclose your score within <probability> and </probability> tags.
* Generate the arguments first then the probability score.
* Generate arguments and probabitlity for each paper separately.
* Do not generate anything else apart from the probability and the arguments.
* Follow this process even if a candidate paper happens to be identical or near-perfect match to the query abstract.

### Response Format for each paper:
<arguments_for>
[Paper ID]: [Reason for including the paper]
Extracted Sentences: "Sentence 1", "Sentence 2", ...
</arguments_for>
<arguments_against>
[Paper ID]: [Reason for not including the paper]
Extracted Sentences: "Sentence 1", "Sentence 2", ...
</arguments_against>
<probability>
[Paper ID]: [Final Probability Score Based on the Arguments]
</probability>

### Your Response:
"""
\end{lstlisting}
    \caption{Prompt used for Debate Ranking.}
    \label{appendix-figure:debate_prompt}
\end{figure*}




\begin{figure}
\centering
\includegraphics[width=\linewidth]{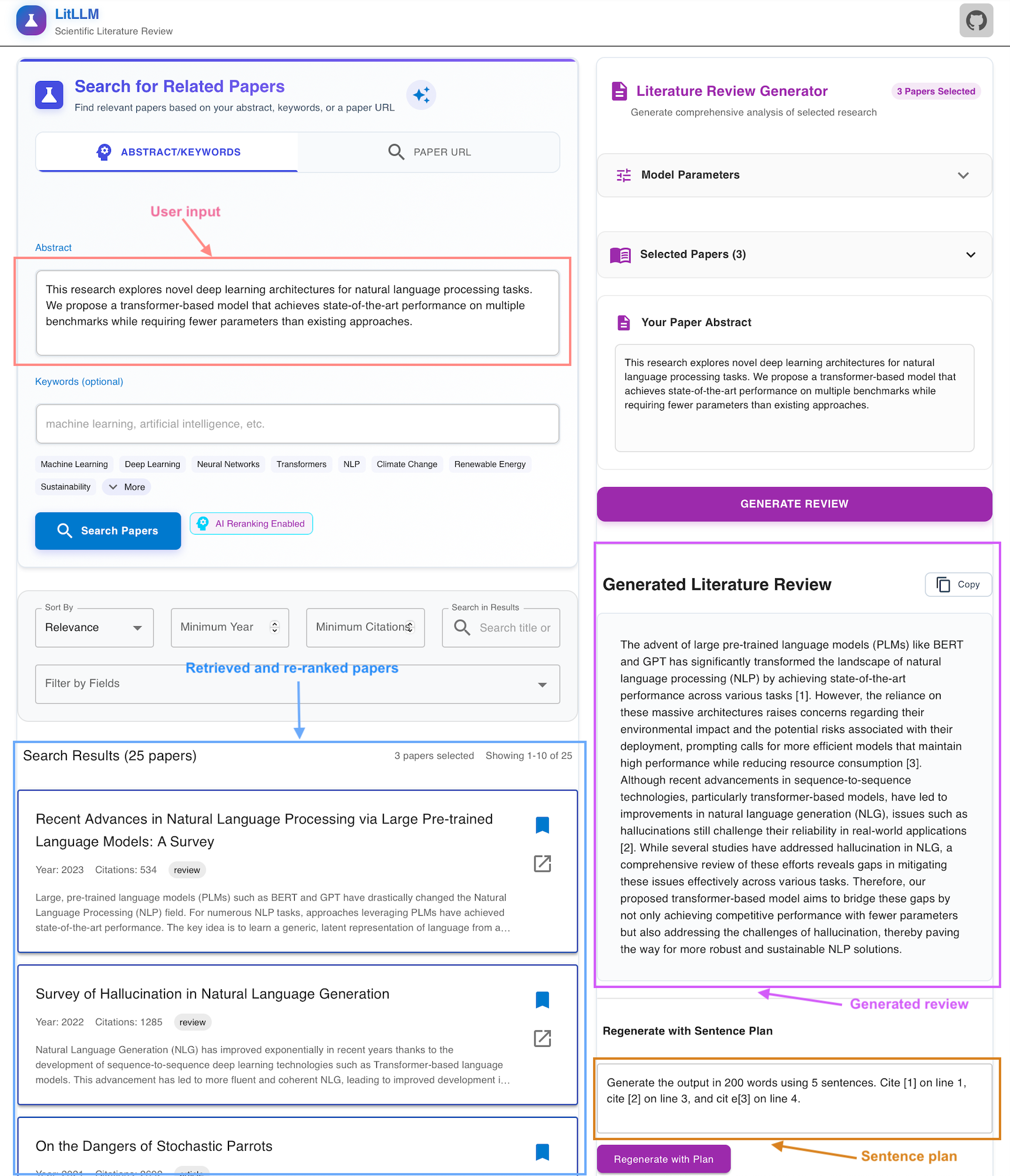}
\caption{LitLLM interface~\citet{agarwal2024litllm}. Our system works on the Retrieval Augmented Generation (RAG) principle to generate the literature review grounded in retrieved relevant papers. The user needs to provide the abstract in the textbox (in purple) and press send to get the generated related work (in red). First, the abstract is summarized into keywords, which are used to query a search engine. Retrieved results are re-ranked (in blue) using an LLM, which is then used as context to generate the related work. Users could also provide a sentence plan (in green) according to their preference to generate a concise, readily usable literature review.}
\label{fig:demo}
\end{figure} 


\end{document}